%% file: main.tex
\pdfoutput=1
\documentclass[10pt,twocolumn,letterpaper]{article}
\usepackage[T1]{fontenc}
\usepackage[pagenumbers]{cvpr}
\makeatletter
\def\@maketitle{\newpage\null\vskip 0.1in\begin{center}{\Large\bf \@title\par}\vspace*{12pt}{\large\lineskip .5em\begin{tabular}[t]{c}\@author\end{tabular}\par}\vspace*{1pt}\end{center}}
\makeatother
\usepackage{amsmath,amssymb,booktabs,graphicx,array,tabularx,longtable}
\usepackage{placeins}
\definecolor{cvprblue}{rgb}{0.21,0.49,0.74}
\usepackage[pagebackref,breaklinks,colorlinks,allcolors=cvprblue]{hyperref}
\hypersetup{pdftitle={EviCraft: Can Minecraft Agents Find What They Cannot See?},pdfauthor={Wenyou Yi, Lantian Zhang, Guanlin Chen, Xingqiu Xu, Haiyang Sun, Minrui Huang, Tianle Liang, Hao Chen, Haorui Xu, Zhou Zhao}}

\input{math_commands.tex}
\makeatletter
\newcommand{\evinum}[2]{\expandafter\def\csname evinum@#1\endcsname{#2}}
\newcommand{\V}[1]{\ifcsname evinum@#1\endcsname\csname evinum@#1\endcsname\else\textbf{[??#1]}\PackageWarning{evinum}{Undefined number #1}\fi}
\makeatother
\input{numbers/numbers.tex}
\title{EviCraft: Can Minecraft Agents Find What They Cannot See?}
\author{%
Wenyou Yi$^{1}$\quad Lantian Zhang$^{2}$\quad Guanlin Chen$^{3}$\quad Xingqiu Xu\quad Haiyang Sun$^{1}$\\
Minrui Huang$^{1}$\quad Tianle Liang$^{1}$\quad Hao Chen$^{4}$\quad Haorui Xu$^{1}$\quad Zhou Zhao$^{1}$\\[0.5em]
{\normalsize $^{1}$Zhejiang University\quad $^{2}$East China University of Science and Technology}\\
{\normalsize $^{3}$University of Wisconsin--Madison\quad $^{4}$Xiamen University}%
}
\begin{document}
\twocolumn[{%
\renewcommand\twocolumn[1][]{#1}%
\maketitle
\centering
\includegraphics[width=\textwidth]{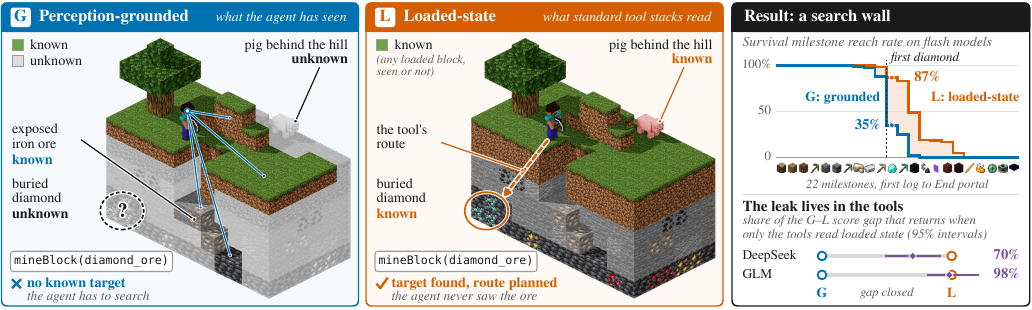}
\captionof{figure}{\textbf{One world, one call, two knowledge states.} Under the perception-grounded regime (G), tools know only what the agent has seen, so \texttt{mineBlock} finds no target; under the loaded-state regime (L) of standard tool stacks, it routes to an ore the agent never saw.}
\label{fig:teaser}
\vspace{7pt}
}]
\begingroup
\renewcommand{\endabstract}{}
\begin{abstract}
\input{sections/00_abstract.tex}
\par
\end{abstract}
\endgroup
\newpage
\vspace*{13pt}
\input{sections/01_introduction.tex}
\input{sections/02_related_work.tex}
\input{sections/03_closure.tex}
\input{sections/04_protocol.tex}
\input{sections/05_experiments.tex}
\input{sections/06_discussion.tex}
\FloatBarrier
\input{sections/07_conclusion.tex}
\input{sections/08_llm_statement.tex}
{\small\bibliographystyle{ieeenat_fullname}\bibliography{references}}
\clearpage
\appendix
\input{sections/A_runtime.tex}
\input{sections/B_protocol.tex}
\input{sections/C_regime_effects.tex}

\input{sections/D_isolated_stages.tex}

\input{sections/E_discovery.tex}
\makeatletter
\def\flushfloatscolumn{\par\begingroup\let\@elt\relax
  \edef\@tempa{\@botlist\@deferlist\@dbldeferlist}%
  \ifx\@tempa\@empty\else\null\newpage\aftergroup\flushfloatscolumn\fi\endgroup}
\makeatother
\flushfloatscolumn
\end{document}

%% file: math_commands.tex
\newcolumntype{L}[1]{>{\raggedright\arraybackslash}p{#1}}

%% file: numbers/numbers.tex
\evinum{budget.diamond.144k}{46\%}
\evinum{budget.diamond.72k}{32\%}
\evinum{budget.gain}{$+$7.4}
\evinum{budget.gain.high}{$+$10.8}
\evinum{budget.gain.low}{$+$4.0}
\evinum{budget.gain.share}{20\%}
\evinum{budget.gain.share.high}{33\%}
\evinum{budget.gain.share.low}{10\%}
\evinum{budget.nf}{36.2}
\evinum{budget.nf.high}{48.2}
\evinum{budget.nf.low}{24.0}
\evinum{budget.nf.share}{97\%}
\evinum{budget.nf.share.high}{117\%}
\evinum{budget.nf.share.low}{75\%}
\evinum{budget.nf.worlds.pos}{10}
\evinum{budget.obsidian.144k}{2\%}
\evinum{budget.residual}{29.8}
\evinum{budget.residual.high}{38.8}
\evinum{budget.residual.low}{21.0}
\evinum{budget.residual.share}{80\%}
\evinum{budget.residual.share.high}{90\%}
\evinum{budget.residual.share.low}{67\%}
\evinum{budget.score.144k}{147.0}
\evinum{budget.score.72k}{139.6}
\evinum{budget.tick.mid}{108,000}
\evinum{budget.ticks}{144,000}
\evinum{budget.worlds.below}{10}
\evinum{combat.fireballruns.G}{17}
\evinum{combat.fireballruns.L}{28}
\evinum{combat.firstfireball.G}{1,331}
\evinum{combat.firstfireball.L}{736}
\evinum{design.reps}{5}
\evinum{design.worlds}{10}
\evinum{disc.G.flash.max}{118.7}
\evinum{disc.G.flash.min}{69.0}
\evinum{disc.ci.high.q9}{$+$4.2}
\evinum{disc.ci.low.q9}{$-$4.5}
\evinum{disc.delta.five.max}{$+$35.1}
\evinum{disc.delta.five.min}{$+$13.9}
\evinum{disc.delta.q9}{$-$0.6}
\evinum{disc.derived.G.ds}{75.5}
\evinum{disc.derived.G.glm}{27.7}
\evinum{disc.diamond.G.ds}{2.0\%}
\evinum{disc.diamond.G.glm}{20.0\%}
\evinum{disc.diamond.L.ds}{0.0\%}
\evinum{disc.diamond.L.glm}{18.0\%}
\evinum{disc.entity.pos.min}{7}
\evinum{disc.flash.gap.derived.max}{98\%}
\evinum{disc.flash.gap.derived.min}{94\%}
\evinum{disc.mineratio.max}{22.8}
\evinum{disc.mineratio.min}{2.2}
\evinum{disc.primary.ci.pos}{4}
\evinum{disc.rel.five.max}{37\%}
\evinum{disc.rel.five.min}{12\%}
\evinum{disc.relgain.derived.max}{$+$20\%}
\evinum{disc.relgain.derived.min}{$-$5\%}
\evinum{disc.relgain.entity.max}{69\%}
\evinum{disc.relgain.entity.min}{52\%}
\evinum{disc.relgain.primary.max}{$+$15\%}
\evinum{disc.relgain.primary.min}{$+$7\%}
\evinum{disc.relgain.subsurf.max}{93\%}
\evinum{disc.relgain.subsurf.min}{27\%}
\evinum{disc.subsurf.G.ds}{11.3}
\evinum{disc.subsurf.G.glm}{12.7}
\evinum{disc.surface.ci.incl0}{4}
\evinum{disc.surface.ci.n}{12}
\evinum{disc.surface.share.G.max}{79\%}
\evinum{disc.surface.share.G.min}{63\%}
\evinum{disc.taxonomy.entity}{102}
\evinum{disc.taxonomy.subsurface}{164}
\evinum{disc.taxonomy.surface_crafted}{218}
\evinum{disc.taxonomy.surface_primary}{69}
\evinum{disc.taxonomy.total}{553}
\evinum{expiry.1.high}{$-$5.0}
\evinum{expiry.1.label}{80 ticks}
\evinum{expiry.1.low}{$-$21.0}
\evinum{expiry.1.shift}{$-$13.6}
\evinum{expiry.2.high}{$+$4.6}
\evinum{expiry.2.label}{2,000 ticks}
\evinum{expiry.2.low}{$-$10.0}
\evinum{expiry.2.shift}{$-$1.8}
\evinum{expiry.3.high}{$+$8.6}
\evinum{expiry.3.label}{no forgetting}
\evinum{expiry.3.low}{$-$6.6}
\evinum{expiry.3.shift}{$+$1.0}
\evinum{expiry.ticks.min}{80}
\evinum{fam.delta.five.max}{$+$30.2}
\evinum{fam.delta.five.min}{$+$12.3}
\evinum{fam.delta.q9}{$+$0.1}
\evinum{fam.excess.G.ds}{13.1}
\evinum{fov.degrees}{110}
\evinum{fov.diamond}{10\%}
\evinum{fov.shift}{$-$12.8}
\evinum{fov.shift.high}{$-$3.0}
\evinum{fov.shift.low}{$-$21.2}
\evinum{harness.abil.G}{119.8}
\evinum{harness.abil.L}{163.4}
\evinum{harness.abil.delta}{$+$43.6}
\evinum{harness.abil.diamond.G}{20}
\evinum{harness.abil.diamond.L}{76}
\evinum{harness.abil.high}{55.4}
\evinum{harness.abil.low}{31.2}
\evinum{harness.abil.nether.G}{0}
\evinum{harness.abil.nether.L}{8}
\evinum{harness.abil.obsidian.G}{0}
\evinum{harness.abil.obsidian.L}{36}
\evinum{harness.abil.worlds.neg}{0}
\evinum{harness.abil.worlds.pos}{10}
\evinum{harness.abil.worlds.tie}{0}
\evinum{harness.delta.max}{$+$55.2}
\evinum{harness.delta.min}{$+$43.6}
\evinum{harness.proto.G}{127.0}
\evinum{harness.proto.L}{182.2}
\evinum{harness.proto.delta}{$+$55.2}
\evinum{harness.proto.diamond.G}{30}
\evinum{harness.proto.diamond.L}{84}
\evinum{harness.proto.high}{67.2}
\evinum{harness.proto.low}{44.6}
\evinum{harness.proto.nether.G}{0}
\evinum{harness.proto.nether.L}{34}
\evinum{harness.proto.obsidian.G}{2}
\evinum{harness.proto.obsidian.L}{52}
\evinum{harness.proto.worlds.neg}{0}
\evinum{harness.proto.worlds.pos}{10}
\evinum{harness.proto.worlds.tie}{0}
\evinum{harness.ref.G}{139.6}
\evinum{harness.ref.L}{176.8}
\evinum{harness.ref.delta}{$+$37.2}
\evinum{harness.ref.diamond.G}{32}
\evinum{harness.ref.diamond.L}{90}
\evinum{harness.ref.high}{44.2}
\evinum{harness.ref.low}{30.8}
\evinum{harness.ref.nether.G}{0}
\evinum{harness.ref.nether.L}{18}
\evinum{harness.ref.obsidian.G}{2}
\evinum{harness.ref.obsidian.L}{50}
\evinum{harness.ref.worlds.neg}{0}
\evinum{harness.ref.worlds.pos}{10}
\evinum{harness.ref.worlds.tie}{0}
\evinum{harness.signflip.p}{0.002}
\evinum{items.ds.G_higher}{152}
\evinum{items.ds.L_higher}{243}
\evinum{items.ds.n}{454}
\evinum{items.ds.neg}{$-$10.0}
\evinum{items.ds.net}{$+$13.9}
\evinum{items.ds.pos}{$+$23.9}
\evinum{items.ds.tied}{59}
\evinum{nocc.diamond.ds}{86\%}
\evinum{nocc.diamond.glm}{76\%}
\evinum{nocc.gap.share.ds}{93\%}
\evinum{nocc.gap.share.glm}{96\%}
\evinum{nocc.gap.share.max}{96\%}
\evinum{nocc.gap.share.min}{93\%}
\evinum{nocc.nether.ds}{14\%}
\evinum{nocc.nether.glm}{14\%}
\evinum{nocc.obsidian.ds}{46\%}
\evinum{nocc.obsidian.glm}{34\%}
\evinum{nocc.score.ds}{174.2}
\evinum{nocc.score.glm}{167.4}
\evinum{nocc.shift.ds}{$+$34.6}
\evinum{nocc.shift.glm}{$+$31.0}
\evinum{nocc.shift.high.ds}{$+$41.0}
\evinum{nocc.shift.high.glm}{$+$36.6}
\evinum{nocc.shift.low.ds}{$+$28.6}
\evinum{nocc.shift.low.glm}{$+$25.2}
\evinum{perm.p.sep.max}{0.025}
\evinum{perm.patterns}{1,024}
\evinum{r32.gap.high}{118\%}
\evinum{r32.gap.low}{78\%}
\evinum{r32.gap.retained}{98\%}
\evinum{reach.diamond.G.flash.max}{40\%}
\evinum{reach.diamond.G.flash.min}{32\%}
\evinum{reach.diamond.G.kimi}{72\%}
\evinum{reach.diamond.L.flash.max}{94\%}
\evinum{reach.diamond.L.flash.min}{76\%}
\evinum{reach.ipick.G.ds}{92\%}
\evinum{reach.ipick.G.flash.max}{92\%}
\evinum{reach.ipick.G.flash.min}{82\%}
\evinum{reach.nether.L.flash.max}{18\%}
\evinum{reach.nether.L.flash.min}{16\%}
\evinum{reach.nether.L.kimi}{52\%}
\evinum{reach.obsidian.G.flash.max}{4\%}
\evinum{reach.obsidian.G.kimi}{12\%}
\evinum{reach.obsidian.L.kimi}{82\%}
\evinum{routecase.G.dist}{466}
\evinum{routecase.L.dist}{35}
\evinum{routecase.L.ticks}{1,286}
\evinum{sensor.low.divisor}{16}
\evinum{sensor.ores.far.recalled}{84}
\evinum{sensor.ores.far.total}{89}
\evinum{sensor.ores.near.recalled}{104}
\evinum{sensor.ores.near.total}{104}
\evinum{sensor.poses}{12}
\evinum{sensor.poses.cave}{6}
\evinum{sensor.poses.surface}{6}
\evinum{sensor.seeds}{3}
\evinum{sensor.still20.far.cave}{97.0\%}
\evinum{sensor.still20.far.surface}{93.5\%}
\evinum{sensor.still20.near.cave}{99.8\%}
\evinum{sensor.still20.near.surface}{98.9\%}
\evinum{sensor.still400.far.cave}{99.3\%}
\evinum{sensor.still400.far.surface}{98.2\%}
\evinum{sensor.walk.far.cave}{98.2\%}
\evinum{sensor.walk.far.low.cave}{88.2\%}
\evinum{sensor.walk.far.low.surface}{64.6\%}
\evinum{sensor.walk.far.surface}{90.7\%}
\evinum{sensor.walk.far.worst}{80.9\%}
\evinum{sensor.walk.near.cave}{99.7\%}
\evinum{sensor.walk.near.surface}{97.9\%}
\evinum{stage.10.G.death}{28}
\evinum{stage.10.G.n}{30}
\evinum{stage.10.G.score}{232.0}
\evinum{stage.10.G.success}{2}
\evinum{stage.10.L.death}{23}
\evinum{stage.10.L.score}{238.3}
\evinum{stage.10.L.success}{7}
\evinum{stage.5.G.n}{30}
\evinum{stage.5.G.success}{15}
\evinum{stage.5.L.success}{30}
\evinum{stage.5.budget}{12,000}
\evinum{stage.6.G.n}{30}
\evinum{stage.6.G.success}{11}
\evinum{stage.6.L.n}{30}
\evinum{stage.6.L.success}{30}
\evinum{stage.6.human.n}{30}
\evinum{stage.6.human.p.G}{0.009}
\evinum{stage.6.human.p.L}{0.005}
\evinum{stage.6.human.success}{22}
\evinum{stage.6.human.timeout}{8}
\evinum{stage.9.G.death}{19}
\evinum{stage.9.G.n}{30}
\evinum{stage.9.L.death}{7}
\evinum{stage.glm.10.G.n}{30}
\evinum{stage.glm.10.G.success}{2}
\evinum{stage.glm.10.L.success}{3}
\evinum{stage.glm.search.G.death}{56}
\evinum{stage.glm.search.G.n}{150}
\evinum{stage.glm.search.G.success}{29}
\evinum{stage.glm.search.G.timeout}{65}
\evinum{stage.glm.search.L.success}{105}
\evinum{stage.nonsearch.G.n}{150}
\evinum{stage.nonsearch.G.success}{148}
\evinum{stage.nonsearch.L.n}{150}
\evinum{stage.nonsearch.L.success}{149}
\evinum{stage.search.G.fail}{109}
\evinum{stage.search.G.n}{150}
\evinum{stage.search.G.success}{41}
\evinum{stage.search.G.timeout}{76}
\evinum{stage.search.L.n}{150}
\evinum{stage.search.L.success}{125}
\evinum{stage6.miny.G}{$-$55}
\evinum{stage6.miny.L}{$-$32}
\evinum{stage6.path.G}{163}
\evinum{stage6.path.L}{66}
\evinum{stage6.pathq1.G}{111}
\evinum{stage6.pathq1.L}{51}
\evinum{stage6.pathq3.G}{263}
\evinum{stage6.pathq3.L}{86}
\evinum{surv.G.flash.range}{3.2}
\evinum{surv.G.kimi}{154.4}
\evinum{surv.L.flash.over.G.kimi.max}{23.4}
\evinum{surv.L.flash.over.G.kimi.min}{14.2}
\evinum{surv.ci.high.kimi}{$+$50.8}
\evinum{surv.ci.high.q9}{$+$23.2}
\evinum{surv.ci.low.kimi}{$+$36.4}
\evinum{surv.ci.low.q9}{$-$0.4}
\evinum{surv.delta.ds}{$+$37.2}
\evinum{surv.delta.five.max}{$+$43.6}
\evinum{surv.delta.five.min}{$+$32.2}
\evinum{surv.delta.flash.max}{$+$40.0}
\evinum{surv.delta.flash.min}{$+$32.2}
\evinum{surv.delta.glm}{$+$32.2}
\evinum{surv.delta.kimi}{$+$43.6}
\evinum{surv.delta.q9}{$+$10.0}
\evinum{surv.kimi.lead.max}{18.0}
\evinum{surv.kimi.lead.min}{14.8}
\evinum{surv.spread.G}{108.2}
\evinum{surv.spread.L}{141.8}
\evinum{ta.A.R.ds}{0.70}
\evinum{ta.A.R.glm}{0.98}
\evinum{ta.A.R.high.ds}{0.92}
\evinum{ta.A.R.high.glm}{1.21}
\evinum{ta.A.R.low.ds}{0.48}
\evinum{ta.A.R.low.glm}{0.81}
\evinum{ta.A.R.max}{98\%}
\evinum{ta.A.R.min}{70\%}
\evinum{ta.A.diamond.ds}{80\%}
\evinum{ta.A.diamond.glm}{88\%}
\evinum{ta.A.nether.glm}{4\%}
\evinum{ta.A.obsidian.glm}{26\%}
\evinum{ta.AT.R.ds}{0.48}
\evinum{ta.AT.R.high.ds}{0.74}
\evinum{ta.AT.R.low.ds}{0.24}
\evinum{ta.G.diamond.ds}{32\%}
\evinum{ta.G.diamond.glm}{32\%}
\evinum{ta.L.diamond.ds}{90\%}
\evinum{ta.L.diamond.glm}{76\%}
\evinum{ta.L.nether.ds}{18\%}
\evinum{ta.L.nether.glm}{16\%}
\evinum{ta.L.obsidian.ds}{50\%}
\evinum{ta.L.obsidian.glm}{38\%}
\evinum{ta.tunnel.calls.ds}{851}
\evinum{ta.tunnel.calls.glm}{547}
\evinum{ta.tunnel.found.before.ds}{737}
\evinum{ta.tunnel.found.before.glm}{394}
\evinum{trans.diamond.gap.max}{65}
\evinum{trans.diamond.gap.min}{27}
\evinum{trans.diamond.held.maxshift}{4.4}
\evinum{trans.remaining.G.max}{59,360}
\evinum{trans.remaining.G.min}{52,474}
\evinum{trans.remaining.L.max}{63,800}
\evinum{trans.remaining.L.min}{63,580}
\evinum{trans.window}{40,000}
\evinum{view.gap.high}{126\%}
\evinum{view.gap.low}{88\%}
\evinum{view.gap.retained}{105\%}
\evinum{view.worlds.above}{10}

%% file: sections/00_abstract.tex
In Minecraft, a common testbed for embodied agents, semantic tools such as \texttt{mineBlock} pick targets from the simulator's loaded world state without a visibility test, so a buried ore the agent has never seen is one call away, and scores cannot show whether the agent searched.
We formalize the missing requirement as \emph{evidence closure}: every decision of an agent system, including those inside its tools, may depend on the world only through facts for which the agent holds evidence.
EviCraft is a runtime and benchmark that enforces closure below the agent and makes the admission rule an experimental variable: a perception-grounded regime (G) admits only what is in line of sight or returned by the agent's own actions, and a loaded-state regime (L) reproduces standard tool stacks, with agents, tools, and worlds held fixed.
On Survival, the regime effect exceeds the strongest agent's lead: under G, three mid-sized agents reach the diamond in \V{reach.diamond.G.flash.min}--\V{reach.diamond.G.flash.max} of runs, against \V{reach.diamond.L.flash.min}--\V{reach.diamond.L.flash.max} under L, and the strongest agent under G scores below each of them under L.
In two models, removing only the line-of-sight test closes \V{nocc.gap.share.min}--\V{nocc.gap.share.max} of the score gap, and loading state only inside tools recovers \V{ta.A.R.min}--\V{ta.A.R.max}, so safeguards on observations and queries leave most of the leak open.
In open-ended acquisition, the gain from L concentrates in items that depend on buried resources or mobs.
Code and benchmark are available at \url{https://github.com/AllenReder/EviCraft}.

%% file: sections/01_introduction.tex
\section{Introduction}
\label{sec:intro}

Finding what is not yet in view is a defining problem of embodied vision, and its benchmarks give the agent only what its sensors deliver from where it stands~\citep{bajcsy1988active,das2018embodied,chaplot2020object,savva2019habitat}.
In Minecraft, progress depends on resources buried in a procedurally generated 3D world~\citep{guss2019minerl,fan2022minedojo}, and language-model agents that act through semantic tools now advance far along its technology chains~\citep{wang2023voyager,qin2024mp5,yu2024mineland,liu2024odyssey,yu2025adam}, yet their scores do not tell us whether the agent searched.
A call such as \texttt{mineBlock(diamond\_ore)} is not a motor command: inside it, a handwritten routine queries the client's loaded chunks, picks a matching block with no line-of-sight test, and walks to it.
The tool is thus part of the agent's policy, and it conditions on simulator state that the agent never perceived.
World facts can reach an agent system at three \emph{read sites}: what the system \emph{pushes} into the model's input, what the model requests through \emph{queries}, and what \emph{tools} read while they execute.
Existing safeguards act on model inputs, the push site and in MP5 also queries~\citep{qin2024mp5,yu2024mineland,yu2025adam}. The tool site stays open~\citep{faldor2026team}, so the interface can solve the search problem the agent appears to solve.

We make this information channel explicit and controllable.
We call an agent system \emph{evidence-closed} when every decision it makes---the content of its observations, the answer to each query, and each choice of target or route inside a tool---depends on the world only through the evidence it holds: the knowledge states that an explicit rule builds, its self-state, and the frames it requests (Section~\ref{sec:closure}).
\textbf{EviCraft} is a Minecraft runtime and benchmark built on this property (Figure~\ref{fig:system}).
The runtime owns the world, the clock, and the knowledge state, and an agent connects through a protocol that has no method for reading the world beyond admitted evidence, so closure holds for any agent confined to it, whatever its prompts or code.
The rule that admits facts then becomes an experimental variable.
The primary \textbf{perception-grounded} regime (G) admits a fact only when it is visible from the agent's eye in the 3D scene or returned by the agent's own actions; its sensing is omnidirectional and its recognition exact, which separates \emph{where to look} from \emph{what is seen}.
A \textbf{loaded-state} regime (L) reproduces the loaded-chunk access of standard tool stacks (Figure~\ref{fig:teaser}).

We study six language models in one reference agent on two tasks whose scores respond differently to a missing fact: \textbf{Survival}, a 22-milestone progression toward End-portal activation in which an unfound resource stalls every later milestone that depends on it, and \textbf{Discovery}, open-order acquisition of distinct item types in which it costs only the items that depend on it.
With worlds and agents held fixed, a difference due to occluded facts should concentrate where they are needed, and it does:
\begin{itemize}
    \item \textbf{Closing the leak exposes a search wall.} Under G, three mid-sized agents stall before the first diamond, a rare and deeply buried resource: they reach it in \V{reach.diamond.G.flash.min}--\V{reach.diamond.G.flash.max} of runs, against \V{reach.diamond.L.flash.min}--\V{reach.diamond.L.flash.max} under L. When one of them runs each stage in isolation, the regimes differ only where a hidden target must be located, and removing only the line-of-sight test from G closes \V{nocc.gap.share.min}--\V{nocc.gap.share.max} of the gap in two of them.
    \item \textbf{More time or a stronger agent does not remove the wall.} With double the budget, one model's G stays below L at the original budget. The strongest agent still stalls just past the diamond under G, with an L--G gap of \V{surv.delta.kimi} points, and under L the mid-sized agents reach the diamond in \V{reach.diamond.L.flash.min}--\V{reach.diamond.L.flash.max} of runs, versus \V{reach.diamond.G.kimi} for the strongest agent under G.
    \item \textbf{The leak lives in the tools.} With pushed observations and model queries grounded, loading state only inside tools recovers \V{ta.A.R.min}--\V{ta.A.R.max} of the gap in two models, so in our runtime, restricting observations and queries leaves most of the leak open.
    \item \textbf{The effect follows dependence on hidden state.} In Discovery, L raises subsurface-dependent items by \V{disc.relgain.subsurf.min}--\V{disc.relgain.subsurf.max} and entity/loot-dependent items by \V{disc.relgain.entity.min}--\V{disc.relgain.entity.max}, while surface drops gain only \V{disc.relgain.primary.min} to \V{disc.relgain.primary.max} and items crafted from surface materials change by \V{disc.relgain.derived.min} to \V{disc.relgain.derived.max}.
\end{itemize}
Our contributions are
(1) \emph{evidence closure} and the three read sites, a criterion for stating what an embodied agent interface knows;
(2) \emph{EviCraft}, a runtime that enforces closure for any agent and exposes the regime per read site, with a two-task benchmark; and
(3) a study locating where agents stall under G and how much of the gap the tool site recovers.

%% file: sections/02_related_work.tex
\section{Related work}
\label{sec:related}

\paragraph{Embodied search and privileged information.}
Embodied question answering and object-goal navigation evaluate whether an agent can find targets from egocentric observations~\citep{das2018embodied,chaplot2020object}, and recent benchmarks score the process of exploration as well as its outcome~\citep{wang2026explore}.
This literature has shown that interface assumptions can decide what a benchmark measures~\citep{anderson2018evaluation}: PointGoal navigation is essentially solved with a perfect GPS and compass~\citep{wijmans2019dd}, and most of the difficulty returns once that sensor is removed~\citep{partsey2022mapping}.
Tool-based language-model agents raise the same issue in a new place: the privileged reader is a tool that stays in the loop at test time, a shortcut~\citep{geirhos2020shortcut} built into the interface.

\paragraph{Minecraft agents: who finds the target?}
Visuomotor policies and world models learn to act from egocentric frames~\citep{guss2019minerl,baker2022video,lifshitz2023steve,hafner2025mastering,hafner2025training}, hierarchical systems pair foundation-model planners with memory, visual grounding, or learned low-level controllers~\citep{wang2023describe,wang2024jarvis,li2024optimus,cai2025rocket,park2025mrsteve}, and recent vision-language-action models learn several action levels in one model~\citep{he2026training,wang2025openha}.
In these systems a learned policy performs the search.
Tool-based agents instead compose semantic APIs into skill libraries~\citep{wang2023voyager,yu2024mineland,liu2024odyssey,yu2025adam,qin2024mp5,faldor2026team}, trading low-level control for long-horizon reasoning, and handwritten code, their helpers or Mineflayer's search and pathfinding, takes over target discovery and navigation by reading world state (Section~\ref{sec:runtime-contract}).
EviCraft keeps the semantic-tool interface that lets language models reach long horizons and controls each read site (Table~\ref{tab:related-comparison}).

\begin{table}[t]
\centering
\caption{\textbf{Read sites in Minecraft agent systems.} \emph{Search}: whether a learned policy or a handwritten tool finds targets. \emph{Input}: how world facts reach the model, excluding self-state and task goals. \emph{Push}, \emph{Query}, \emph{Tool}: whether, in released code, the model input, the model's requests, and target search and path planning inside tools are limited to perceivable state (\checkmark: in code; --: not limited, or only by a prompt instruction; n/a: no such site). \emph{Regimes}: whether the knowledge rule can be varied with the agent fixed.}
\label{tab:related-comparison}
\scriptsize
\setlength{\tabcolsep}{2.1pt}
\begin{tabular}{@{}lcccccc@{}}
\toprule
 & & & \multicolumn{3}{c}{Read site grounded} & \\
\cmidrule(lr){4-6}
System & Search & Input & Push & Query & Tool & Regimes \\
\midrule
MineRL~\citep{guss2019minerl}, VPT~\citep{baker2022video} & Policy & RGB & \checkmark & n/a & n/a & -- \\
JARVIS-1~\citep{wang2024jarvis}, Optimus-1~\citep{li2024optimus} & Policy & RGB & \checkmark & n/a & n/a & -- \\
Voyager~\citep{wang2023voyager} & Tool & Text & -- & -- & -- & -- \\
Odyssey~\citep{liu2024odyssey} & Tool & Text & -- & n/a & -- & -- \\
MP5~\citep{qin2024mp5} & Tool & RGB+text & \checkmark & \checkmark & -- & -- \\
ADAM~\citep{yu2025adam} & Tool & RGB+text & \checkmark & n/a & -- & -- \\
MineLand~\citep{yu2024mineland} & Tool & RGB+text & \checkmark & -- & -- & -- \\
\midrule
\textbf{EviCraft (ours)} & Tool & RGB+text & \checkmark & \checkmark & \checkmark & per site \\
\bottomrule
\end{tabular}
\end{table}

\paragraph{Safeguards and evaluation.}
Embodied benchmarks for language-model agents supply symbolic state, object coordinates, or automatic navigation~\citep{li2024embodied,yang2025embodiedbench}, and recent harnesses outside Minecraft build scene graphs only from observations and verified action outcomes~\citep{wang2026towards} or gate task state by visibility against a hidden backend~\citep{wang2026compiling}.
Within Minecraft, MP5 adds active ego-view perception~\citep{qin2024mp5}, ADAM removes metadata from planner inputs~\citep{yu2025adam}, and MineLand limits its camera's view distance and tells the model to look up only blocks in view~\citep{yu2024mineland}; these measures govern what the model sees or asks, while the released tools still select targets from world state.
MineNPC-Task bounds perception and action by loaded chunks as a proxy for line of sight~\citep{doss2026minenpc}, a bound that still admits occluded blocks; Section~\ref{sec:anatomy} opens only the tool site to measure its share of the gap.
Minecraft evaluations measure serial depth along technology milestones~\citep{guss2019minerl,wang2024jarvis,li2024optimus}, breadth as the number of distinct items~\citep{wang2023voyager,liu2024odyssey}, or compositional tasks~\citep{zheng2023mcu,ju2026mineexplorer}, and progress metrics reveal what binary success hides~\citep{ma2024agentboard}.

%% file: sections/03_closure.tex
\section{Evidence-closed agent interfaces}
\label{sec:closure}
\label{sec:runtime}

\begin{figure*}[t]
\centering
\includegraphics[width=\textwidth]{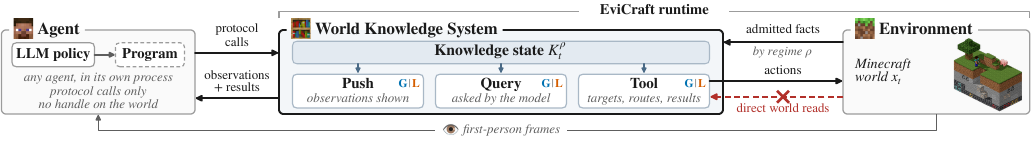}
\caption{\textbf{Closure enforced below the agent.} The three read sites read one knowledge state $K_t^\rho$, which the environment updates only with facts that the regime $\rho$ admits; the regime can be set per site (G\,$|$\,L). The direct world read of standard semantic tools (red) does not exist.}
\label{fig:system}
\end{figure*}

\subsection{Tools are part of the policy}
\label{sec:runtime-contract}
Let $x_t$ denote the full simulator state at tick $t$.
A language-model agent does not act on $x_t$ directly: the model emits \emph{calls} (queries, actions, or programs that compose them), and an \emph{interface} executes each call as a multi-step routine that selects targets, plans routes, and issues motor commands.
The system that acts is the composition of the two, and its information is whatever either part reads.

In standard stacks the interface reads $x_t$.
Mineflayer-based~\citep{prismarine2023mineflayer} collection helpers~\citep{wang2023voyager,yu2024mineland,liu2024odyssey,yu2025adam} locate targets with \texttt{bot.findBlocks}, which searches the client's loaded chunks with no line-of-sight test, and model-written code in Voyager and MineLand can call it directly.
MP5's released controller~\citep{qin2024mp5} picks movement targets from MineDojo's voxel observation~\citep{fan2022minedojo}, a block grid copied from the world with no occlusion filter.
World facts therefore enter at three read sites: \emph{push} (the prompt or camera), \emph{query} (a call the model makes to ask about the world), and \emph{tool} (reads made while an action executes, and whatever the action then reports back).
The composite policy is a function of $(h_t,x_t)$, not of the observation history $h_t$ alone~\citep{kaelbling1998planning}: a privileged policy in the vocabulary of imitation learning~\citep{chen2020learning}, here kept in the loop at test time. Its score thus measures the agent plus what its tools read, not search from observations.

\subsection{Evidence closure}
\label{sec:closure-def}
We replace the implicit read of $x_t$ by an explicit knowledge state.
A \emph{regime} $\rho$ is an admission rule: from each world event $\xi_t$ (a change in what is visible, a contact, the outcome of an action) it updates a set of records,
\begin{equation}
K_t^\rho=\mathcal U_\rho\bigl(K_{t-1}^\rho,\xi_t\bigr),
\label{eq:update}
\end{equation}
where a record holds the state of a block or entity as last admitted and the tick of admission.
A cell is a known block, known air, or \emph{unknown}~\citep{hornung2013octomap}: absence of evidence is represented, not defaulted, and tools must plan around it.
The agent's self-state (position, health, inventory) is always available, and the model may request first-person frames, which show only surfaces visible from the eye (Section~\ref{sec:constrained-execution}).
We call the knowledge trajectory $K_{0:t}^\rho$, the self-states, and the requested frames up to $t$ the \emph{evidence history} $E_{0:t}^\rho$.

\paragraph{Definition (evidence closure).}
An agent system is \emph{evidence-closed} under $\rho$ if the distribution of each of its decisions---the observation assembled for the model, the result of each query, each target and route chosen inside a tool, and the result that the tool returns---depends on the world only through $E_{0:t}^\rho$.
Equivalently, for the same agent, two worlds $x$ and $x'$ that induce the same evidence history up to $t$ yield the same distribution for each decision at $t$.

The second form is a non-interference condition~\citep{goguen1982security}: a fact without evidence cannot influence behavior.
Closure constrains decisions, not dynamics. Collision, damage, and block breaking run on $x_t$. Action feedback enters $K^\rho$ under the regime's admission rule before it can affect a target, route, or return value; choices inside tools read external world facts only from $K^\rho$ (Appendix~\ref{app:runtime-contract}).
Throughout, a \emph{leak} means a violation of closure under G; a system under L is closed but leaks in this sense. Every leak has a read site.

\paragraph{Regimes per site.}
Each read site can carry its own rule, so a configuration is a triple $(\rho_{\mathrm{push}},\rho_{\mathrm{query}},\rho_{\mathrm{tool}})$.
Closure under $\rho$ is $(\rho,\rho,\rho)$; grounding push and query but retaining loaded-state tools gives $(\mathrm G,\mathrm G,\mathrm L)$.
Changing $\rho$ at all sites with the agent, tools, and world fixed measures an empirical value of privileged information for that agent; changing it at one site locates where that value enters.

\subsection{Two regimes}
\label{sec:regimes}

\paragraph{Perception-grounded (G).}
G admits an external fact on two kinds of evidence, sight and contact.
At every tick, rays cast from the eye over the full sphere admit the first occluding surface within 64 blocks along each direction, the air in front of it, and each entity that a ray reaches; a block that the agent breaks, places, touches, or finds to contradict a record is updated at once (Appendix~\ref{app:runtime}).
Recognition is exact: an admitted block carries its true type.
While walking, it records \V{sensor.walk.near.surface} (surface) and \V{sensor.walk.near.cave} (caves) of the blocks that a denser reference sensor sees within 32 blocks (Appendix~\ref{app:runtime}).
G is thus an idealized visual sensor: omnidirectional, independent of lighting, and with exact type labels, it is limited only by occlusion and by this sampling recall.
A record is dropped after 400 ticks unobserved, so G is perception with a short memory; Section~\ref{sec:anatomy} varies this window.

\paragraph{Loaded-state (L).}
L reproduces the semantics of a Mineflayer client's chunk cache, the state that standard tool stacks read: $K_t^{\mathrm L}$ contains every block and entity in the chunk columns that the server currently delivers to the agent (four chunks in each direction, over the full height of the world), regardless of occlusion.
Records follow the world and are evicted when a column leaves the delivery boundary; beyond it L knows nothing (Appendix~\ref{app:runtime}).
G and L thus differ in occlusion, range, and retention (Section~\ref{sec:anatomy}).

\subsection{Enforcing closure at the protocol}
\label{sec:constrained-execution}
A rule stated in a prompt can be ignored, and a rule implemented in an agent's own code must be audited again for every agent. EviCraft places the rule below the agent (Figure~\ref{fig:system}).
The \emph{runtime}, a server-side mod, owns the avatar, the world, the clock, and $K_t^\rho$. The agent under test is a separate process that holds the model, its prompts, and its program executor, and speaks a small protocol: refresh knowledge from current perception, query $K_t^\rho$, call a tool, wait, and capture a first-person frame.
All tools execute inside the runtime, where (i) queries return only records of $K_t^\rho$; (ii) composite actions, such as collecting a block type or tunneling until one is found, choose targets among those records; and (iii) path planning reads a snapshot of $K_t^\rho$ and replans as contact and sight update it during the action.
The one input that does not pass through $K_t^\rho$ is the first-person frame, rendered from the avatar's eye out to the range of G's rays and L's delivery boundary. It shows only surfaces visible from the current pose, evidence of the kind that sight admits to $K_t^{\mathrm G}$, and enters the evidence history directly; renderer and range are the same under every regime.
Ground truth for scoring flows through a separate channel for which the agent holds no credential.
Closure composes: if the agent's initialization and seeds are independent of the initial world, any program that reaches the world only through closed calls is closed (Appendix~\ref{app:runtime}), so model-written programs and third-party agents inherit it.

The runtime also owns time: the world advances only while a tool call executes or the avatar is dead, so inference latency costs no game time.

\subsection{A leak test from counterfactual worlds}
\label{sec:boundary}
Composition reduces closure for any agent to closure of the runtime's own calls, so a test need only probe those calls, at queries, target selection, route planning, and entity tracking (Appendix~\ref{app:runtime}).
In ordinary play the records and the world mostly agree, so the test builds pairs of worlds in which they come apart and drives the protocol with the same deterministic script in both.
In a \emph{hidden-difference} pair the worlds differ in a fact never admitted under G, so a G-closed interface must act identically in both: an ore sealed behind a wall at one of two sites yields no record and is mined only in a seen-first control.
In a \emph{false-belief} pair a fact admitted under G changes out of view, so a G-closed interface must act on its record and be wrong in one world: a route cell seen as air and replaced by stone stays air in the planner's first snapshot, and an interaction with a moved animal heads for its last seen position and fails.
Across these seven cases, under L the same script mines the sealed ore, plans with the stone, and reaches the moved animal, which shows that each hidden fact is one that a tool reads; three further cases check correction, forgetting, and coverage.

%% file: sections/04_protocol.tex
\section{Tasks and evaluation protocol}
\label{sec:protocol}
\subsection{Tasks}
\label{sec:shared-protocol}
Both tasks run in Minecraft 1.21.1 survival mode on easy difficulty, with an empty initial inventory and respawning enabled.
Budgets are $T_S=72{,}000$ game ticks (one hour of game time) for Survival and $T_D=144{,}000$ for Discovery, counted by the runtime clock.
A G and an L configuration differ in one field, the regime.
An offline scorer receives the true world state every 20 ticks (Appendix~\ref{app:protocol}).

\paragraph{Survival: progression depth.}
\label{sec:survival-protocol}
The agent is asked to progress toward activating an End portal.
We score 22 milestones with fixed values $v_j\in[10,300]$ in four phases, each of which needs items from the one before (Appendix~\ref{app:protocol}): \emph{basic resources} that are visible from the surface (wood, stone, a furnace; 10--70); \emph{underground resources} that are out of sight (iron, then diamond, which is rare and deep, then obsidian, which forms where lava meets water; 90--180); the \emph{Nether}, a hostile dimension entered through an obsidian portal, where a fortress must be found and blazes killed (190--260); and the \emph{End}, which requires locating a buried stronghold (280--300).
With $\tau_j$ the tick at which milestone $j$ is first confirmed ($\infty$ if never),
\begin{equation}
\begin{aligned}
S(\tau)&=\max\bigl(\{0\}\cup\{v_j:\tau_j\le\tau\}\bigr),\\
\mathrm{nAUC}_S&=\tfrac{1}{300\,T_S}\textstyle\int_0^{T_S}S(\tau)\,d\tau.
\end{aligned}
\label{eq:survival-curve}
\end{equation}
The endpoint $S(T_S)$ gives the deepest tier reached, and $\mathrm{nAUC}_S$ also rewards reaching it early.
Because point values are set by design, we also report the fraction of runs that \emph{reach} each milestone, that is, end with $S(T_S)\ge v_j$.

\paragraph{Discovery: acquisition breadth.}
\label{sec:discovery-protocol}
The agent is asked to obtain as many distinct item types as possible, in any order, by gathering, mining, smelting, hunting, and crafting.
With $I_k$ the set of item IDs held at scoring sample $k$ (tick $\tau_k$),
\begin{equation}
D(\tau)=\Bigl|\bigcup_{k:\tau_k\le\tau}I_k\Bigr|,\qquad
\overline D=\frac{1}{T_D}\int_0^{T_D}\!D(\tau)\,d\tau.
\label{eq:discovery-curve}
\end{equation}
Consumed or lost items stay counted. Item IDs distinguish wood and color variants; breadth with variants merged into families gives the same conclusions (Appendix~\ref{app:discovery-diagnostics}).

\subsection{Agents}
\label{sec:eval-setup}
We evaluate six language models in one reference agent: DeepSeek v4.1 Flash, Qwen 3.8 Flash, and GLM 5.3 Flash, which we call the \emph{Flash models}; the smaller Qwen 3.8 27B and Qwen 3.5 9B; and the larger Kimi K3.1 Pro.
The reference agent gives every model the same prompt, ability library built on the runtime's tools, and orchestration: at each turn the model writes a program that calls abilities, and it can request first-person frames (Appendix~\ref{app:runtime}).

\subsection{World-paired estimation}
\label{sec:paired}
Each task uses $W=\V{design.worlds}$ initial worlds, with \V{design.reps} repetitions per model, world, and regime from identical snapshots.
For a metric $Y$ we average repetitions within each world and regime to obtain $\bar Y_{w,\rho}$, and report the matched-world effect
\begin{equation}
\widehat\Delta_{\mathrm{L}-\mathrm{G}}(Y)=\frac{1}{W}\sum_{w=1}^{W}\bigl(\bar Y_{w,\mathrm L}-\bar Y_{w,\mathrm G}\bigr).
\label{eq:paired-regime-effect}
\end{equation}
Every comparison, between regimes or between models, uses this matched-world difference with a 95\% percentile bootstrap interval over worlds; two configurations are \emph{separated} when the interval excludes zero.

\subsection{What a hidden-state account implies}
\label{sec:implications}
If $\widehat\Delta_{\mathrm{L}-\mathrm{G}}$ reflects access to occluded facts and not some other difference between the conditions, four observations should follow.
\textbf{(I1)} On Survival, the loss under G concentrates at milestones whose target is occluded: runs that reach the prerequisite of such a milestone cross it less often in the same time, and a doubled budget leaves most of the loss.
\textbf{(I2)} When each stage of the chain is run in isolation, with its prerequisites supplied, the regimes differ where a hidden target must be located and agree elsewhere.
\textbf{(I3)} Removing occlusion from G or opening the tool site alone recovers a large part of the gap, whereas reducing L's delivery range or removing G's block forgetting leaves the gap intact.
\textbf{(I4)} On Discovery, the gain from L concentrates in items that depend on occluded facts and is small for surface drops and of mixed sign for items crafted from them.

%% file: sections/05_experiments.tex
\section{Experiments}
\label{sec:results}
\input{tables/main-score/main-results.tex}
\input{figures/regime-effects/latex_include.tex}

\subsection{The regime changes scores and comparisons}
\label{sec:benchmark-separation}
\label{sec:regime-effects}
Table~\ref{tab:main-score} gives all scores and Figure~\ref{fig:regime-effects} their progression.
Switching from G to L raises the mean Survival endpoint of every model. For the three Flash models the effect is \V{surv.delta.flash.min} to \V{surv.delta.flash.max} points, and for Kimi K3.1 Pro it is \V{surv.delta.kimi}, both positive in all \V{design.worlds} worlds.
L also raises Discovery breadth by \V{disc.delta.five.min} to \V{disc.delta.five.max} item types, for every model except Qwen 3.5 9B (\V{disc.delta.q9} [\V{disc.ci.low.q9}, \V{disc.ci.high.q9}]).
The time-integrated $\mathrm{nAUC}_S$ and $\overline D$ move in the same direction as the endpoints for every model.
In a second, general-purpose tool-calling harness, the effect is \V{harness.delta.min} to \V{harness.delta.max} points (Appendix~\ref{app:regime-effects}).

\paragraph{The regime effect exceeds the strongest agent's lead.}
Under G, Kimi K3.1 Pro leads each Flash model by \V{surv.kimi.lead.min}--\V{surv.kimi.lead.max} Survival points. Under L, each Flash model scores \V{surv.L.flash.over.G.kimi.min}--\V{surv.L.flash.over.G.kimi.max} points above the strongest agent's G score of \V{surv.G.kimi}.
Reach rates point the same way: a Flash model under L reaches the diamond in \V{reach.diamond.L.flash.min}--\V{reach.diamond.L.flash.max} of runs, and the strongest agent under G in \V{reach.diamond.G.kimi}.

\paragraph{The weakest agent gains least.}
Qwen 3.5 9B gains \V{surv.delta.q9} Survival points, with an interval [\V{surv.ci.low.q9}, \V{surv.ci.high.q9}] that lies below every other model's; the other five gain \V{surv.delta.five.min} to \V{surv.delta.five.max}.
Known targets help only an agent that can complete the tool chain that exploits them, so the spread of Survival scores grows from \V{surv.spread.G} points under G to \V{surv.spread.L} under L.

\subsection{A wall at hidden resources}
\label{sec:stage-diagnostics}
\input{figures/search-wall/latex_include_main.tex}
Under G the Flash models end Survival within \V{surv.G.flash.range} points of one another, and all three milestone curves drop at the diamond (I1; Figure~\ref{fig:wall}).
Under both regimes nearly every Flash run smelts iron, which lies underground but is common and often exposed in caves.
Under G the reach rate then falls from \V{reach.ipick.G.flash.min}--\V{reach.ipick.G.flash.max} at the iron pickaxe to \V{reach.diamond.G.flash.min}--\V{reach.diamond.G.flash.max} at the diamond and at most \V{reach.obsidian.G.flash.max} at obsidian; under L, \V{reach.diamond.L.flash.min}--\V{reach.diamond.L.flash.max} reach the diamond and \V{reach.nether.L.flash.min}--\V{reach.nether.L.flash.max} the Nether.

\paragraph{The wall persists with time to spare.}
Under G the median Flash run reaches the iron pickaxe with \V{trans.remaining.G.min}--\V{trans.remaining.G.max} of its 72{,}000 ticks left, against \V{trans.remaining.L.min}--\V{trans.remaining.L.max} under L.
From that point, the estimated probability of reaching the diamond within \V{trans.window} ticks is lower under G in each of the five models with enough such runs, by \V{trans.diamond.gap.min} to \V{trans.diamond.gap.max} percentage points (Appendix~\ref{app:regime-effects}).
More time helps slowly: doubling G's budget to \V{budget.ticks} ticks raises the score of DeepSeek v4.1 Flash by \V{budget.gain} points [\V{budget.gain.low}, \V{budget.gain.high}] and its diamond rate from \V{budget.diamond.72k} to \V{budget.diamond.144k}, while obsidian stays at \V{budget.obsidian.144k}, no run enters the Nether, and the score stays \V{budget.residual} points [\V{budget.residual.low}, \V{budget.residual.high}] below L at half the budget (Figure~\ref{fig:anatomy}(b)).

\paragraph{A stronger agent moves the wall.}
Kimi K3.1 Pro reaches the diamond in \V{reach.diamond.G.kimi} of G runs, against \V{reach.diamond.G.flash.min}--\V{reach.diamond.G.flash.max} for the Flash models, and stalls just past it (thin lines in Figure~\ref{fig:wall}): \V{reach.obsidian.G.kimi} reach obsidian and none enter the Nether, against \V{reach.obsidian.L.kimi} and \V{reach.nether.L.kimi} under L.
Its own L--G gap remains \V{surv.delta.kimi} points [\V{surv.ci.low.kimi}, \V{surv.ci.high.kimi}].

\input{figures/gap-anatomy/latex_include.tex}

\subsection{In isolated stages, the gap follows hidden targets}
\label{sec:stages}
To separate the difficulty of a milestone from delays inherited from earlier ones (I2), we cut the chain into eleven stages and run DeepSeek v4.1 Flash on each in isolation: a new agent starts in one world at a prescribed position with the stage's prerequisites, objective, and budget (Figure~\ref{fig:anatomy}(a); Appendix~\ref{app:survival-stages}).
Stages that need no hidden target---gathering and crafting from resources visible on the surface (Stages 1--4) and building a portal from supplied obsidian (Stage 8)---succeed in \V{stage.nonsearch.G.success} of \V{stage.nonsearch.G.n} executions under G and \V{stage.nonsearch.L.success} of \V{stage.nonsearch.L.n} under L: loaded state adds almost nothing where nothing must be found.
The five stages that require locating iron, diamonds, obsidian, a fortress, or the End portal (Stages 5--7, 9, and 11) succeed in \V{stage.search.G.success} of \V{stage.search.G.n} executions under G and \V{stage.search.L.success} of \V{stage.search.L.n} under L.
GLM 5.3 Flash shows the same pattern on the search stages (\V{stage.glm.search.G.success} and \V{stage.glm.search.L.success} of \V{stage.glm.search.G.n}; Appendix~\ref{app:survival-stages}).

\paragraph{Search costs time or lives.}
Most of G's \V{stage.search.G.fail} failures in the search stages are timeouts (\V{stage.search.G.timeout}). In the diamond stage the median G execution travels \V{stage6.path.G} blocks, against \V{stage6.path.L} under L (Appendix~\ref{app:survival-stages}).
Thirty human operators who call the same abilities and read the same $K_t^{\mathrm G}$, rendered in 3D instead of as records, complete it in \V{stage.6.human.success} of \V{stage.6.human.n} executions, against \V{stage.6.G.success} for the model under G and \V{stage.6.L.success} under L, so the diamond can be found within the budget from what G admits (Appendix~\ref{app:survival-stages}).
In the fortress stage \V{stage.9.G.death} of \V{stage.9.G.n} G executions end in death (\V{stage.9.L.death} under L), consistent with longer exposure while searching.
Iron shows this cost at a scale that a full run absorbs: G completes its stage (\V{stage.5.budget} ticks, no respawning) in \V{stage.5.G.success} of \V{stage.5.G.n} executions against \V{stage.5.L.success} under L, yet \V{reach.ipick.G.ds} of the model's full G runs reach the iron pickaxe.

\paragraph{Beyond search.}
In Stage 10, which starts armed in a fortress, known targets are not enough: completion rises only from \V{stage.10.G.success} to \V{stage.10.L.success} of \V{stage.10.G.n} executions; the rest die.

\subsection{Anatomy of the gap}
\label{sec:anatomy}
G and L differ in occlusion, range, and retention. Figure~\ref{fig:anatomy}(b) varies each factor and then the tool site alone (I3); $R=(\bar Y_{\mathrm{abl}}-\bar Y_{\mathrm G})/(\bar Y_{\mathrm L}-\bar Y_{\mathrm G})$ is the share of the gap closed.

\paragraph{Occlusion carries the gap; range and retention do not.}
G$^\circ$ removes only G's line-of-sight test and keeps its range and retention. DeepSeek v4.1 Flash then gains \V{nocc.shift.ds} points [\V{nocc.shift.low.ds}, \V{nocc.shift.high.ds}] over G, \V{nocc.gap.share.ds} of the way to L, and the diamond in \V{nocc.diamond.ds} of runs (G: \V{ta.G.diamond.ds}; L: \V{ta.L.diamond.ds}); GLM 5.3 Flash gains \V{nocc.shift.glm} [\V{nocc.shift.low.glm}, \V{nocc.shift.high.glm}], \V{nocc.gap.share.glm} of its gap.
Reducing L's delivery boundary from four chunks to two retains \V{view.gap.retained} of the gap, and capping L at 32 blocks of the eye, the lookup range of Mineflayer-based tools such as Voyager's, retains \V{r32.gap.retained}. Shortening G's block-retention window to \V{expiry.ticks.min} ticks moves its score by \V{expiry.1.shift} points [\V{expiry.1.low}, \V{expiry.1.high}], while longer windows up to no forgetting move it by \V{expiry.2.shift} and \V{expiry.3.shift}, with intervals including zero; without block forgetting L still leads by \V{budget.nf} points [\V{budget.nf.low}, \V{budget.nf.high}].
Narrowing G's sight to a \V{fov.degrees}-degree cone around the look direction, which covers the first-person frame, moves its score by \V{fov.shift} points [\V{fov.shift.low}, \V{fov.shift.high}] and the diamond to \V{fov.diamond} of runs, so omnidirectional G understates the gap for a camera-limited agent.

\paragraph{The tool site alone recovers most of the gap.}
In the \emph{tools-alone} configuration $(\mathrm G,\mathrm G,\mathrm L)$, pushed observations and the model's queries stay grounded, and only the reads inside tools, including what tools report back, see loaded state.
The agent nevertheless recovers $R={}$\V{ta.A.R.ds} [\V{ta.A.R.low.ds}, \V{ta.A.R.high.ds}] for DeepSeek v4.1 Flash, and \V{ta.A.R.glm} [\V{ta.A.R.low.glm}, \V{ta.A.R.high.glm}] for GLM 5.3 Flash, an interval that includes full recovery.
Recovery is largest at the diamond (G, tools-alone, L: \V{ta.G.diamond.ds}, \V{ta.A.diamond.ds}, \V{ta.L.diamond.ds} for DeepSeek v4.1 Flash; \V{ta.G.diamond.glm}, \V{ta.A.diamond.glm}, \V{ta.L.diamond.glm} for GLM 5.3 Flash), and both stay below L at obsidian (Appendix~\ref{app:regime-effects}).
One tool shows the mechanism. Of the \V{ta.tunnel.calls.ds} tunneling calls by DeepSeek v4.1 Flash, \V{ta.tunnel.found.before.ds} return the coordinates of a target before breaking any block (GLM 5.3 Flash: \V{ta.tunnel.found.before.glm} of \V{ta.tunnel.calls.glm}); grounding that one lookup lowers $R$ to \V{ta.AT.R.ds}.

\input{figures/discovery-diagnostics/latex_include_main.tex}

\subsection{In Discovery, the gain follows hidden state}
\label{sec:graded}
In Discovery the hidden-state account implies a selective gain and not a wall (I4).
We group every item by its least knowledge-dependent route, using loot and rare drops only where other routes fail (Appendix~\ref{app:discovery-diagnostics}): \emph{surface-derived} items are made from surface materials, \emph{surface-primary} items drop from natural surface blocks, \emph{entity/loot-dependent} items need a mob or loot, and \emph{subsurface-dependent} items need an ore, a deep-layer block, or a Nether or End material.
The gain from L falls in the two dependent categories (Figure~\ref{fig:discovery-effects}). Relative to G, L raises subsurface-dependent breadth by \V{disc.relgain.subsurf.min}--\V{disc.relgain.subsurf.max} and entity/loot-dependent breadth by \V{disc.relgain.entity.min}--\V{disc.relgain.entity.max}, and changes surface-primary breadth by \V{disc.relgain.primary.min} to \V{disc.relgain.primary.max} and surface-derived breadth by \V{disc.relgain.derived.min} to \V{disc.relgain.derived.max} (ratios exclude Qwen 3.5 9B, which holds almost no such items).
The intervals of the two dependent categories exclude zero for every model except Qwen 3.5 9B; surface-primary gains are smaller, though their intervals exclude zero for \V{disc.primary.ci.pos} models, and surface-derived ones change sign.
G agents also issue \V{disc.mineratio.min}--\V{disc.mineratio.max} times as many blind digging actions as L agents.
With surface items at \V{disc.surface.share.G.min}--\V{disc.surface.share.G.max} of G breadth, total breadth rises only \V{disc.rel.five.min}--\V{disc.rel.five.max}.
Under G, Discovery also separates every pair of Flash models, which Survival does not (\V{disc.G.flash.min} to \V{disc.G.flash.max} item types), mostly through surface-derived items (\V{disc.flash.gap.derived.min}--\V{disc.flash.gap.derived.max} of each gap).

%% file: tables/main-score/main-results.tex
\begin{table*}[t]
\centering\small
\caption{\textbf{Scores under perception-grounded (G) and loaded-state (L) knowledge.} Equal-world means over ten matched worlds; $\Delta_{\mathrm{L}-\mathrm{G}}$ is the matched-world endpoint effect of Eq.~\ref{eq:paired-regime-effect} with its 95\% world-bootstrap interval (all effects: Appendix~\ref{app:regime-effects}).}
\label{tab:main-score}
\setlength{\tabcolsep}{3.1pt}
\begin{tabular}{@{}l ccc cc @{\hspace{10pt}} ccc cc@{}}
\toprule
& \multicolumn{5}{c}{\textbf{Survival} (progression depth)} & \multicolumn{5}{c}{\textbf{Discovery} (acquisition breadth)}\\
\cmidrule(lr{10pt}){2-6}\cmidrule(l){7-11}
& \multicolumn{3}{c}{Endpoint $S(T_S)$} & \multicolumn{2}{c}{$\mathrm{nAUC}_S$} & \multicolumn{3}{c}{Endpoint $D(T_D)$} & \multicolumn{2}{c}{Time average $\overline D$}\\
\cmidrule(lr){2-4}\cmidrule(lr{10pt}){5-6}\cmidrule(lr){7-9}\cmidrule(l){10-11}
Model & G & L & $\Delta_{\mathrm{L}-\mathrm{G}}$ & G & L & G & L & $\Delta_{\mathrm{L}-\mathrm{G}}$ & G & L\\
\midrule
Kimi K3.1 Pro & $154.4$ & $198.0$ & $+43.6$\,{\scriptsize[36.4, 50.8]} & $.431$ & $.538$ & $136.8$ & $171.9$ & $+35.1$\,{\scriptsize[29.9, 40.5]} & $82.0$ & $105.3$\\
DeepSeek v4.1 Flash & $139.6$ & $176.8$ & $+37.2$\,{\scriptsize[30.8, 44.2]} & $.398$ & $.509$ & $118.7$ & $132.6$ & $+13.9$\,{\scriptsize[5.7, 23.3]} & $81.9$ & $90.8$\\
Qwen 3.8 Flash & $137.8$ & $177.8$ & $+40.0$\,{\scriptsize[33.0, 46.6]} & $.392$ & $.509$ & $93.8$ & $113.6$ & $+19.8$\,{\scriptsize[14.4, 25.8]} & $64.1$ & $76.8$\\
GLM 5.3 Flash & $136.4$ & $168.6$ & $+32.2$\,{\scriptsize[25.4, 40.0]} & $.382$ & $.483$ & $69.0$ & $83.5$ & $+14.5$\,{\scriptsize[8.4, 20.8]} & $45.8$ & $58.0$\\
Qwen 3.8 27B & $122.0$ & $157.8$ & $+35.8$\,{\scriptsize[24.6, 49.6]} & $.351$ & $.439$ & $54.5$ & $74.5$ & $+20.0$\,{\scriptsize[16.1, 24.2]} & $41.1$ & $52.4$\\
Qwen 3.5 9B & $46.2$ & $56.2$ & $+10.0$\,{\scriptsize[-0.4, 23.2]} & $.138$ & $.152$ & $19.4$ & $18.8$ & $-0.6$\,{\scriptsize[-4.5, 4.2]} & $14.4$ & $13.0$\\
\bottomrule
\end{tabular}
\end{table*}

%% file: figures/regime-effects/latex_include.tex
\begin{figure*}[t]
\centering
\includegraphics[width=\textwidth]{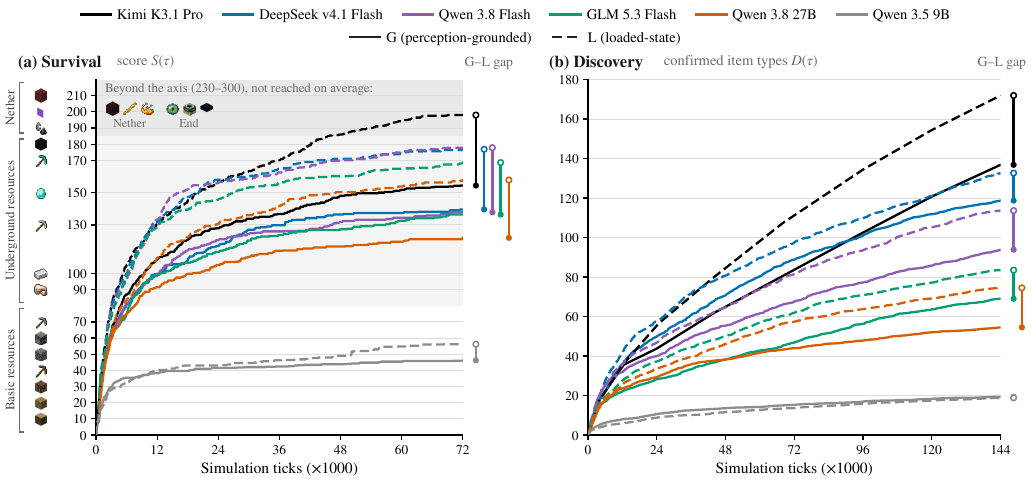}
\caption{\textbf{Progression over time} (equal-world means) on (a) Survival and (b) Discovery. Lines at the right span each model's G (filled) and L (hollow) endpoints; in (a), icons mark the rubric milestones of Section~\ref{sec:survival-protocol}, grouped by phase.}
\label{fig:regime-effects}
\end{figure*}

%% file: figures/search-wall/latex_include_main.tex
\begin{figure}[t]
\centering
\includegraphics[width=\columnwidth]{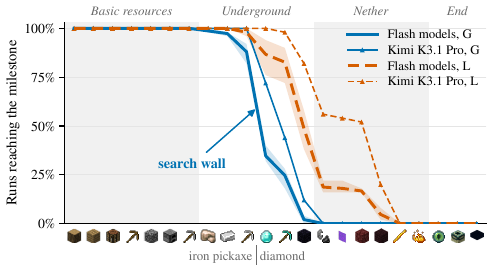}
\caption{\textbf{Reach along the Survival chain.} Share of runs that reach each milestone within $T_S$. Flash models: mean (line) and range (band). Every model: Appendix~\ref{app:regime-effects}.}
\label{fig:wall}
\end{figure}

%% file: figures/gap-anatomy/latex_include.tex
\begin{figure*}[t]
\centering
\includegraphics[width=\textwidth]{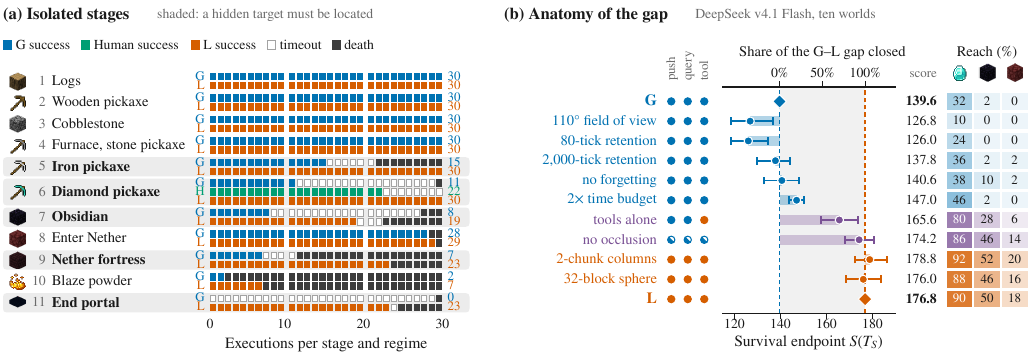}
\caption{\textbf{Locating and decomposing the gap} (DeepSeek v4.1 Flash). \textbf{(a)} Eleven Survival stages of one world, each run in isolation; one cell per execution. \textbf{(b)} Survival endpoint under interventions: the regime at each read site (dots; half-filled: no occlusion), the shift from the unmodified regime (bars, 95\% intervals), and the share of runs reaching the diamond, obsidian, and the Nether.}
\label{fig:anatomy}
\end{figure*}

%% file: figures/discovery-diagnostics/latex_include_main.tex
\begin{figure}[t]
\centering
\includegraphics[width=\columnwidth]{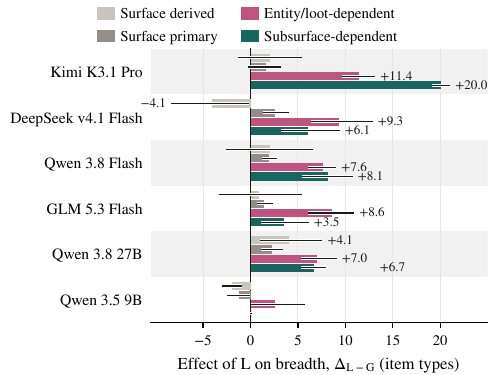}
\caption{\textbf{The gain from L falls on items that depend on hidden state.} L$-$G effect on Discovery breadth per category, with 95\% intervals; effects with $|\Delta|\geq 3$ are labeled.}
\label{fig:discovery-effects}
\end{figure}

%% file: sections/06_discussion.tex
\section{Discussion}
\label{sec:discussion}

\paragraph{Where the wall comes from.}
Under G, every admitted block carries its true type, and the agent's skill library states at which depths ores generate and prescribes branch mining for diamonds; the agents act on it, digging blindly several times more often and traveling farther than under L.
With occlusion alone closing most of the gap and range, retention, and budget each leaving most of it, the likely bottleneck is searching an occluded 3D volume under time and hazards, which loaded-state tools bypass and the stronger model did not overcome. For these agents, this is an open problem of active 3D prospecting: like frontier exploration~\citep{yamauchi1997frontier} and next-best-view planning~\citep{bircher2016receding}, it decides where to look next, but exposes surfaces by digging, as in mechanical search~\citep{danielczuk2019mechanical}, and must decide when to leave a region.
The wall arises under exact, omnidirectional recognition, so it is a problem of where to look and not of what is seen. G's idealizations (all-around sight, no lighting, exact labels) favor G; the one we tested, a camera field of view, widens the gap (Section~\ref{sec:anatomy}), and a detector on the requested frames would add recognition error under G only; we leave it as a further controlled variable.

\paragraph{Limitations.}
The model comparison uses the program-writing reference harness, without the curriculum and growing skill library of Voyager~\citep{wang2023voyager} or the planner--actor--critic split of Odyssey~\citep{liu2024odyssey}. Mineflayer-based agents cannot run unchanged, as their helpers execute in a client that receives the loaded chunks; L reproduces their access instead. Other control designs and action abstractions may search better under G or shift the balance between search and combat failures.
The isolated stages use one world and two agents.

%% file: sections/07_conclusion.tex
\section{Conclusion}
\label{sec:conclusion}
EviCraft enforces evidence closure below the agent, inside tools as well as in the model.
With the leak closed, agents stall at buried resources; the loss follows hidden targets; in two models, lifting occlusion alone recovers nearly all of it and the tool site alone most of it; and a stronger model moves the wall while retaining a large regime gap.
In our evaluation, loaded-state tools find the hidden targets that decide these scores, so benchmarks should report the admission rule at each read site.
Searching an occluded 3D world remains open for the agents we tested; EviCraft makes it measurable.

%% file: sections/08_llm_statement.tex
\section*{LLM Usage Statement}
Claude Code and claude-opus-5-5 were used to assist with code refinement.
Code review, experimental design, and result analysis were performed by the human authors.
claude-opus-5-5 was also used to polish the manuscript’s wording and optimize its LaTeX layout; all resulting changes were checked by the authors.
No generative image tools were used to directly create the paper’s figures or charts.

%% file: sections/A_runtime.tex
\section{Runtime, regimes, and reference agent}
\label{app:runtime}

\subsection{Architecture and knowledge acquisition}
\label{app:runtime-contract}

\paragraph{Three components, two interfaces.}
The \emph{runtime} is a server-side mod that owns the avatar, the world, game time, the knowledge state, and the facts that scoring reads.
The \emph{agent} (a harness: model client, prompts, program executor) is a separate process that connects over a local JSON-RPC protocol with one credential.
The \emph{benchmark runner} starts the server, prepares the world, starts the agent, and records the scoring samples through a second, authoritative protocol for which the agent holds no credential.
The agent-facing protocol has methods to refresh knowledge from current perception, to query knowledge (single blocks, blocks of a type, block regions, entities), to list and call tools, to wait a number of ticks, and to capture a first-person frame.
Its knowledge queries never read the live world, and the tool catalogue is the same for every agent.
The regime is fixed per world at initialization and both regimes expose the same self-state, queries, and tools; a G and an L configuration differ in that one field.

\paragraph{Game time.}
The runtime counts the budget and stops the world on the budget tick.
The world advances only while a tool call is in progress, including an explicit wait, or while the avatar is dead or respawning; queries, observation refreshes, and the gaps between two actions of one program cost no game time.
The agent is told neither the length of the budget nor the point values.

\paragraph{Perception-grounded regime (G).}
Blocks enter $K_t^{\mathrm G}$ through sight rays cast from the eye and through contact with the body.
The ray directions of one pass form a regular equi-angular cube-map lattice over the full sphere, $32\times32$ directions per face ($6{,}144$ rays); each pass shifts the lattice by the next point of a low-discrepancy sequence, so successive passes interleave into finer coverage. A tick spends a fixed work budget of $2\times10^6$ units, one per cell a ray steps through plus two per ray, and carries an unfinished pass forward; rays reach 64 blocks. A direction whose last ray and whose neighbors' last rays met no block (open sky) is cast on one pass in eight. Once the eye and the terrain in view have been still for 1{,}024 passes, a tick casts a single pass.
Glass, ice, water, and plants that can be walked through are seen without ending a ray; every other block ends the ray where its outline shape is hit, so leaves and lava hide what is behind them.
An entity within 64 blocks is admitted when a ray reaches any of ten sample points of its bounding box (eye, center, and eight inset corners); spectators and entities invisible to the avatar are excluded.
Blocks that the agent modifies, breaks, or finds to differ from its records are updated immediately.
Block and entity records are dropped 400 ticks (20\,s) after their last observation; Appendix~\ref{app:regime-effects} varies the block window.
Appendix~\ref{app:boundary-validation} measures how much of what is visible this sampling records.

\paragraph{Loaded-state regime (L).}
$K_t^{\mathrm L}$ follows the semantics of a Mineflayer client's chunk cache without running such a client: every cell, with air recorded as known air, and every entity in the chunk columns that the server currently delivers to the avatar, over the full height of the dimension and with no test of sight or distance.
View and simulation distances are four chunks, so block knowledge reaches at least 64 blocks horizontally. Changes inside delivered columns update records in place, and when a column leaves the delivery boundary or the avatar changes dimension its records are removed.

\paragraph{Grounded regime without occlusion (G$^\circ$).}
G$^\circ$ is used only in the ablation of Section~\ref{sec:anatomy}. It differs from G in one rule: a ray that hits a block or an entity continues instead of ending, so nothing occludes what lies behind it. Every other rule of G is unchanged: the ray lattice, the work budget, the 64-block range, the admission of entities through their sample points, the immediate update of blocks that the agent modifies, breaks, or finds to differ from its records, and the 400-tick windows of block and entity records. All three read sites read $K_t^{\mathrm G^\circ}$.

\paragraph{Grounded regime with a field of view.}
This variant is also used only in the ablation of Section~\ref{sec:anatomy}. It differs from G in one rule: rays are cast only within a cone of \V{fov.degrees} degrees full angle around the look direction, the cone that circumscribes the $854{\times}480$ first-person frame at the camera's 70-degree vertical field of view. When the look direction changes, the sensor resumes full passes in the new direction. Every other rule of G is unchanged, and all three read sites read the resulting knowledge state.

\paragraph{Reads of the world during execution.}
The path planner reads a snapshot of $K_t^\rho$; it may route through unknown cells, stops before the first unknown cell, and replans from new evidence. A tool's choices at tick $t$ therefore depend on $x$ only through the records admitted up to $t$, and its return value is assembled from the same records and the self-state.
Table~\ref{tab:runtime-reads} lists, for every method of the protocol and every tool in the catalogue, what the call reads to choose a target, a route, or its result, and what it reads of the live world while the body executes.

\paragraph{Proposition (closure under composition).}
Fix a regime $\rho$ and an agent whose internal state before receiving initial evidence, together with all decision-making random seeds, is independent of the initial world $x$. Suppose that (a) the agent's program receives world information only through protocol observations and returns, and calls execute sequentially; (b) for fixed call arguments and seeds, each observation, query result, and choice of target or route inside a tool depends on $x$ only through the evidence available before that decision, and each tool's return value only through the evidence available before it returns; and (c) world events update records only through $K_t^\rho=\mathcal U_\rho(K_{t-1}^\rho,\xi_t)$, while self-state and requested frames enter $E_{0:t}^\rho$ directly. Then every decision of the program depends on $x$ only through $E_{0:t}^\rho$. With the same initial state and seeds, the agent's decisions in two worlds coincide for as long as the two evidence histories coincide. The distribution in the definition averages this function over initialization and seeds at a fixed evidence history; it is not the distribution conditioned on observing that history in natural runs, where the seeds that produce a given history may differ between worlds.

\paragraph{Proof sketch.}
Order all decisions and evidence updates, within and across calls, including those at the same tick. Fix the initial state and seeds. Each program step is a function of its internal state and the protocol returns received so far, which by (b) and (c) are functions of the evidence prefix. Physical outcomes may depend on $x_t$, but they reach a later decision only through an admission step. By induction, each decision is a function of the evidence prefix, the initial internal state, and the seeds, with no further dependence on $x$. For any fixed evidence history, averaging this function over the same world-independent law of initialization and seeds gives the same decision distribution in both worlds.

\begin{table*}[t]
\centering
\caption{The calls of the agent-facing protocol and what each reads. \emph{Decision reads}: what a call uses to choose a target, a route, or its result. Every world fact among them is a record of $K_t^\rho$ or part of the self-state, and every call that moves the body plans its route over a snapshot of $K_t^\rho$. A menu that the avatar's own action has opened, the time of day, and the difficulty enter the evidence history with the self-state and are the same under every regime. \emph{Server-side execution}: the reads of the live world that a physical action needs to run and to verify its outcome. \emph{Check}: the bound block, cell, or entity still matches its record, and the native ray from the eye reaches it. A verified outcome, and a target found to differ from its record, are written to $K_t^\rho$ by the regime's rule (Section~\ref{sec:regimes}); nothing else these reads see enters $K_t^\rho$, so they leak no world knowledge into it. A dash marks no read. The remaining methods (\texttt{runtime.info}, \texttt{actor.get}, \texttt{actor.create}, \texttt{tool.list}, \texttt{task.cancel}, \texttt{runtime.time.await\_tick}) read no world facts. The composite abilities of the reference agent (Table~\ref{tab:composites}) reach the world only through these calls.}
\label{tab:runtime-reads}
\scriptsize
\begin{tabularx}{\textwidth}{@{}>{\raggedright\arraybackslash}p{0.25\textwidth}>{\raggedright\arraybackslash}X>{\raggedright\arraybackslash}p{0.29\textwidth}@{}}
\toprule
Call & Decision reads & Server-side execution \\
\midrule
\multicolumn{3}{@{}l}{\emph{Perception and knowledge}} \\
\texttt{observation.observe} & -- & admission by the regime's rule \\
\texttt{knowledge.query} & $K_t^\rho$ & -- \\
\texttt{observation.capture\_view} & -- & a frame rendered at the eye \\
\midrule
\multicolumn{3}{@{}l}{\emph{Status and lookup, without game time}} \\
\texttt{get\_self\_status} & health, hunger, position, dimension, equipment, effects, inventory & -- \\
\texttt{get\_world\_info} & time of day, difficulty & -- \\
\texttt{lookup\_registry}, \texttt{lookup\_recipe} & game registries and recipes & -- \\
\texttt{manage\_scaffold\_materials} & the block patterns that navigation may place & -- \\
\texttt{inspect\_gui} & the open menu, or the inventory menu & -- \\
\midrule
\multicolumn{3}{@{}l}{\emph{Travel}} \\
\texttt{goto} & the goal; for a block goal, the easiest reachable match in $K_t^\rho$ & movement \\
\texttt{goto\_block\_interaction\_pose} & a known block; standable poses around it & movement; check from each pose reached \\
\texttt{goto\_block\_placement\_pose} & a target cell and its known supports; standable poses around it & movement; check from each pose reached \\
\midrule
\multicolumn{3}{@{}l}{\emph{Block actions, without travel}} \\
\texttt{break\_block} & a known block; the best carried tool for its type & check; breaking; drops \\
\texttt{activate\_block} & a known block & check; opened menu or changed state \\
\texttt{use\_item\_on\_block} & a known block, or a known fluid source for an empty bucket; a carried item & check; changed item and blocks \\
\texttt{place\_block} & known air in every cell the block fills; a known support & check; placed blocks; consumed item \\
\texttt{enter\_bed} & a known bed & check; the game's rules for sleeping \\
\texttt{use\_item} & a known air cell to aim at, with known air up to the eyes & native use; changed item \\
\midrule
\multicolumn{3}{@{}l}{\emph{Entity actions}} \\
\texttt{interact\_entity} & a tracked entity by UUID, at its last known position & check on the entity \\
\texttt{attack} & tracked entities by UUID, or the hostile ones visible in $K_t^\rho$ & hits; the targets' movement; drops \\
\texttt{collect\_items} & tracked drops by UUID or item pattern, near their known positions & pickup \\
\midrule
\multicolumn{3}{@{}l}{\emph{Inventory and menus}} \\
\texttt{craft} & recipes, the inventory, the open crafting menu & the menu \\
\texttt{transfer}, \texttt{menu\_button}, \texttt{close\_gui} & the open menu & the menu \\
\texttt{equip\_item}, \texttt{eat}, \texttt{drop\_items} & the inventory & the dropped item (\texttt{drop\_items}) \\
\texttt{throw\_eye\_of\_ender} & the inventory & the eye's flight over five ticks \\
\texttt{wait} & -- & game time passes \\
\bottomrule
\end{tabularx}
\end{table*}
\paragraph{Reference agent.}
The reference agent builds 46 abilities on the runtime's tools: thirty single-step abilities (movement, block and entity interaction, crafting, inventory, and knowledge queries) and sixteen composites written against the same protocol (Table~\ref{tab:composites}).
A composite reaches the world only through the knowledge queries and tools available to the model's own program, so it reads $K_t^\rho$ wherever it selects a target or plans a route.
At each turn the model receives its self-state, the known blocks around the body, and the known entities, and replies with a TypeScript program that calls abilities; the program runs in an isolated context without access to the process, the file system, or module loading.
The model can also load one of seven handwritten skills (early tools, ore mining, smelting and crafting, combat, Nether portal, fortress and blazes, End portal), keep a plan, and request a first-person frame of $854{\times}480$ pixels rendered by a camera client at the avatar's eye; the three most recent frames stay in the request.
All six evaluated models accept image input, and each frame is attached to the next request as an image.
Frames do not pass through $K_t^\rho$, and the renderer and its range are the same under G and L.

\begin{table*}[!t]
\centering
\caption{The sixteen composite abilities of the reference agent and the knowledge each reads to select targets. A lookup by block type ranks the matching records of $K_t^\rho$ by distance and, apart from the caps listed, searches all of $K_t^\rho$, which under L spans the delivered chunk columns over the full height of the world. Every composite also plans its movements over a snapshot of $K_t^\rho$; under L these reads see loaded state, and in the tools-alone configuration they read $K_t^{\mathrm L}$.}
\label{tab:composites}
\scriptsize
\begin{tabular}{@{}ll@{}}
\toprule
Composite & What it selects from $K_t^\rho$ \\
\midrule
\multicolumn{2}{@{}l}{\emph{Lookup by block type}} \\
\texttt{collect\_blocks} & up to 32 nearest remembered blocks matching a type pattern, or supplied coordinates \\
\texttt{tunnel\_for\_blocks} & nearest remembered match before digging; after each step, the cells around the body ($5{\times}4{\times}5$), \\
 & and the remembered matches again once new air is known \\
\texttt{craft} & nearest remembered crafting table within 32 blocks, else known free cells to place one \\
\texttt{smelt\_at\_furnace} & nearest remembered furnace within 32 blocks \\
\texttt{fill\_bucket} & up to four nearest remembered water or lava source cells \\
\texttt{sleep\_in\_bed} & remembered beds within 32 blocks, or a supplied bed \\
\midrule
\multicolumn{2}{@{}l}{\emph{Local regions around the body}} \\
\texttt{explore\_and\_return} & exploration anchors scored by unknown cells in a radius of 4 blocks, up to 128 blocks from the start by default \\
\texttt{mine\_down} & the cells below the feet and around the next step of a staircase \\
\texttt{look\_around} & a $17{\times}17$ height map of known columns around the body \\
\texttt{place\_block\_nearby} & known air or replaceable cells in a $5{\times}5$ set over three layers around the feet \\
\texttt{suffocation\_escape} & adjacent cells with known support \\
\texttt{surface} & the column above the body while underwater \\
\midrule
\multicolumn{2}{@{}l}{\emph{Exact cells supplied by the model}} \\
\texttt{goto\_and\_break\_block} & interaction pose for a known block, and the blocks occluding it \\
\texttt{goto\_and\_activate\_block} & interaction pose for a known block \\
\texttt{goto\_and\_place\_block} & placement pose for a target cell \\
\texttt{goto\_and\_use\_item\_on\_block} & interaction pose for a known block \\
\bottomrule
\end{tabular}
\end{table*}

\paragraph{Tools-alone configuration.}
The configuration $(\mathrm G,\mathrm G,\mathrm L)$ of Section~\ref{sec:anatomy} maintains both knowledge states with the same rules.
The automatically assembled observation and the queries that the model's program calls directly read $K_t^{\mathrm G}$.
Everything else reads $K_t^{\mathrm L}$: sub-calls of composite actions, target selection, path-planning snapshots, interaction checks, and the values that tools return. The model's own reads therefore stay within what it has perceived, but a tool it invokes can find hidden targets, plan over unseen terrain, and report what it found.
This access pattern resembles systems that restrict the planner's input while the tools in their released code read world state~\citep{qin2024mp5,yu2025adam}, although tools and ranges differ.
In particular, the tunneling action's lookup of remembered targets and its inspection of the cells ahead read $K_t^{\mathrm L}$, so a tunnel can end at a hidden ore without digging toward it; the sensitivity variant of Table~\ref{tab:tools-alone} grounds only the lookup.
Two scripted cases confirm this access pattern (Appendix~\ref{app:boundary-validation}).

\subsection{Models, worlds, and instructions}
\label{app:settings}
\paragraph{Models.}
Table~\ref{tab:model-settings} lists the model configurations. The self-hosted models are served with SGLang 0.5.22 on H200 nodes; Kimi K3.1 Pro is queried through a commercial hosted API. Requests set only the reasoning setting; each model samples with the parameters recommended in its model card, which the server applies by default. Every configuration has a context window of $262{,}144$ tokens and at most $65{,}536$ output tokens per request.

\begin{table*}[!t]
\centering
\caption{Model configurations of the reference agent. A dash: no top-$k$ truncation.}
\label{tab:model-settings}
\footnotesize
\begin{tabular}{@{}llll@{}}
\toprule
Model & Checkpoint or API model & Reasoning setting & Temperature / top-$p$ / top-$k$ \\
\midrule
Kimi K3.1 Pro & \texttt{moonshotai/Kimi-K3.1-Pro} & high & 1.0 / 0.95 / -- \\
DeepSeek v4.1 Flash & \texttt{deepseek-ai/DeepSeek-V4.1-Flash-1017} & high & 1.0 / 1.0 / -- \\
Qwen 3.8 Flash & \texttt{Qwen/Qwen3.8-Flash-Next} & xhigh & 1.0 / 0.95 / 20 \\
GLM 5.3 Flash & \texttt{zai-org/GLM-5.3-Flash} & high & 1.0 / 0.95 / -- \\
Qwen 3.8 27B & \texttt{Qwen/Qwen3.8-27B} & xhigh & 1.0 / 0.95 / 20 \\
Qwen 3.5 9B & \texttt{Qwen/Qwen3.5-9B} & template default (thinking on) & 1.0 / 0.95 / 20 \\
\bottomrule
\end{tabular}
\end{table*}

\paragraph{Worlds.}
Both tasks use frozen Minecraft 1.21.1 snapshots of seeds 100--109, each with a fixed spawn. The spawn stands on solid ground with clear headroom, full health and food, and an empty inventory. Five spawns were adjusted before freezing, moved inland from a shore or lowered from a tree canopy.

\paragraph{Task instructions.}
The model receives the task as a fixed assignment in its prompt. The Survival assignment reads: ``Starting with an empty inventory, advance as far as you can through Minecraft's survival progression. The key milestones, in order, are: obtain iron ingots, activate a Nether portal, and activate an End portal. Progress is judged by the furthest milestone you have ever reached; it is not lost when items are later used up, dropped, or lost on death. The run lasts a fixed amount of game time and ends automatically; you cannot end it early, so keep working toward the next milestone until it stops. If you die, you respawn and continue. Plan your own exploration and actions, and use legitimate in-game methods.'' The Discovery assignment reads: ``Starting with an empty inventory, obtain as many different item types as you can. An item type counts once, the first time you hold it in your inventory, a hand, or an armor slot. It still counts after you use it up, drop it, store it, or lose it on death, so you do not need to keep items, fill your inventory, or collect many of one type. Item types are distinguished by item ID: for example, oak planks and birch planks are different types. The run lasts a fixed amount of game time and ends automatically; you cannot end it early, so keep finding new item types until it stops. If you die, you respawn and continue. Choose your own objectives and use legitimate in-game methods.''

\subsection{Read sites in released code}
\label{app:code-audit}
Table~\ref{tab:code-audit} lists the commits we read for the tool-based rows of Table~\ref{tab:related-comparison} and the call through which each system's tools select targets; we read this code and did not run it.
In the Mineflayer-based systems this call is \texttt{bot.findBlocks} or \texttt{bot.findBlock}, which iterates over the blocks of the client's loaded chunks and has no line-of-sight test; its body is identical in Mineflayer 4.8.1 and 4.14.0, the versions these systems allow or pin.
MP5 states that its human-designed interface code is not released, so its Tool entry rests on the released structured actions, which locate mining and approach targets in the voxel grid that MineDojo copies from the world around the agent.
MP5's query site is grounded because, in its default active-perception mode, the planner obtains scene information only by asking a multimodal perceptor about the current ego-view image.

\begin{table*}[!t]
\centering
\caption{Commits behind the tool-site entries of Table~\ref{tab:related-comparison} and the call through which each system's tools select targets. \emph{Range}: \texttt{maxDistance} in blocks; for MP5, the half-width of the voxel grid.}
\label{tab:code-audit}
\scriptsize
\setlength{\tabcolsep}{4pt}
\begin{tabular}{@{}lllr@{}}
\toprule
System & Repository @ commit & Target selection & Range \\
\midrule
Voyager & \texttt{MineDojo/Voyager@55e45a8} & \texttt{bot.findBlocks} & 32 \\
Odyssey & \texttt{zju-vipa/Odyssey@260303c} & \texttt{bot.findBlocks} & 32 \\
ADAM & \texttt{OpenCausaLab/ADAM@959dccd} & \texttt{bot.findBlock} & 32 \\
MineLand & \texttt{cocacola-lab/MineLand@f72d3f2} & \texttt{bot.findBlocks} & 48 \\
MP5 & \texttt{IranQin/MP5@deb7bd8} & MineDojo voxel grid & 5 \\
\bottomrule
\end{tabular}
\end{table*}

\subsection{Validating closure and the regimes}
\label{app:boundary-validation}
\input{tables/boundary-validation/observed-outcomes.tex}
A difference between G and L measures access to hidden facts only if the runtime does what Section~\ref{sec:closure} states: no decision reads the world except through the evidence history; $K_t^\rho$ holds what the rule $\rho$ admits, for as long as the rule says; and G's sensor records what is in view.
This section checks the three in turn.

\paragraph{What the architecture gives.}
The agent is a separate process whose only connection to the world is the protocol (Appendix~\ref{app:runtime-contract}), so it can affect the Minecraft environment only through protocol calls; by the composition proposition, its programs are closed whenever the runtime's calls are.
The leak test checks the implementation against this construction at the points where a query answers or a tool chooses a target or a route.

\paragraph{Scripted cases.}
The test has twelve cases, each a scripted scene in a fresh peaceful superflat world.
The first ten run once under G and once under L with the runtime configuration of the main experiments (Table~\ref{tab:boundary-full}); the last two run in the tools-alone configuration.
The script issues its requests through the knowledge and action interfaces that agents use, and no language model takes part.
A separate authoritative channel builds the scene, changes it out of view, places the body where a case prescribes, and reads blocks, inventories, entities, and delivered chunks afterwards; it writes nothing into $K_t^\rho$.
The expected outcome of each run is an assertion in the script, fixed before the run, and every run meets it.

\paragraph{Hidden difference: a sealed ore (three cases).}
The bot stands in a closed cobblestone room with walls one block thick and holds an iron pickaxe.
A diamond ore lies outside the room at one of two sites, neither of which the bot has seen, and the two worlds differ only in the site.
The script queries for diamond ore and then calls the collection action.
Under G the query returns no record and nothing is mined, at either site.
In the control, the bot first visits the ore and sees it, and the same call then mines it.
Under L the ore is a record in all three cases and is mined.
In the case at the first site and in the control, the query also passes through the model-facing harness with a scripted stand-in for the model, and its result reaches the model's request unchanged.

\paragraph{False belief: a changed route cell and a moved animal (four cases).}
\emph{Route cell.} The bot sees air at a cell on the route to a navigation goal. An opaque wall is then built in front of the cell, and behind it the cell is left as air or replaced by stone.
We read the state of the cell in the first snapshot that the planner takes, before the body reaches the cell, and cancel the navigation.
G's snapshot holds air in both worlds; L's holds air in one and stone in the other.
Where the cell was replaced, L's planner thus starts from a snapshot that holds the stone; this planner input is what Section~\ref{sec:boundary} refers to as planning with the stone.
\emph{Animal.} The bot sees a sheep. A wall then hides it, and the sheep stays in place or is moved six blocks into a sealed room.
The script asks for the sheep's position and calls the interaction that shears it.
Under G the record holds the old position in both worlds; where the sheep was moved, the interaction ends with a target-lost failure and the sheep is not sheared.
Under L the record follows the sheep, and the interaction shears it in both worlds.

\paragraph{Correction, forgetting, and coverage (three cases).}
These cases check the rules of Section~\ref{sec:regimes}: how a record is corrected, when it is dropped, and what each regime covers.
When the bot sees the moved sheep again, G's record takes the new position and the interaction succeeds.
When the moved sheep instead stays out of view for 500 ticks, G drops its record as the 400-tick window ends, while L holds the new position throughout; seeing the sheep again restores G's record.
In the coverage case the body is placed far from a column that holds a diamond block on the surface and a second one buried below it, then near the column, then far again, with the game clock stopped.
L holds both blocks while the column is delivered, but neither before nor after.
G holds only the surface block, from the moment it is in view, and keeps it after leaving, since distance does not remove a record.
The surface block is then changed to gold while the body is far away: G keeps the diamond record, L has none, and a return visit updates both to gold.

\paragraph{Separate control of the read sites (two cases).}
The last two cases run in the tools-alone configuration $(\mathrm G,\mathrm G,\mathrm L)$ and in its variant with the tunneling lookup grounded (Appendix~\ref{app:runtime-contract}).
A diamond ore lies behind a wall within the radius of the pushed observation.
In both cases the ore is absent from the pushed observation, from the block queries that the model's program calls directly, and from the first model request, while the collection action and block-mode navigation reach it.
The tunneling action returns the coordinates of the ore without digging, and starts to dig once its lookup is grounded.

\input{figures/sensor-recall/latex_include.tex}

\paragraph{Recall of the block sensor.}
The leak test checks that the probed calls do not read beyond admitted records. The converse is not automatic: G admits a block only when a ray reaches it, so sampling could leave visible blocks unrecorded and make search harder than visibility alone would.
We measure this against a reference sensor that spends four times the work per tick on a different lattice, at \V{sensor.poses} poses (\V{sensor.poses.surface} on the surface and \V{sensor.poses.cave} in caves) of \V{sensor.seeds} Survival worlds, with no model and no task (Figure~\ref{fig:sensor-recall}).
Standing still for one second, the sensor as configured in the experiments holds \V{sensor.still20.near.surface} (surface) and \V{sensor.still20.near.cave} (caves) of the reference blocks within 32 blocks, and \V{sensor.still20.far.surface} and \V{sensor.still20.far.cave} of those at 32--64 blocks; after 400 ticks the latter reach \V{sensor.still400.far.surface} and \V{sensor.still400.far.cave}.
After 400 ticks of walking it holds \V{sensor.walk.near.surface} and \V{sensor.walk.near.cave} of the reference memory within 32 blocks, and \V{sensor.walk.far.surface} and \V{sensor.walk.far.cave} beyond (\V{sensor.walk.far.worst} at the worst pose).
Of the ore blocks that the reference first sees within 32 blocks, the sensor records \V{sensor.ores.near.recalled} of \V{sensor.ores.near.total}, and \V{sensor.ores.far.recalled} of \V{sensor.ores.far.total} of those first seen farther away.
With 1/\V{sensor.low.divisor} of the work budget, recall while standing converges to the same level more slowly, and recall at 32--64 blocks while walking falls to \V{sensor.walk.far.low.surface} and \V{sensor.walk.far.low.cave}. At the budget used in the experiments the curve is nearly flat underground, where the search for buried resources takes place, and still rising for open terrain beyond 32 blocks, so a denser sensor would add distant surface blocks and little else.
Recall is measured against this denser sampler rather than an analytic visibility test.

%% file: tables/boundary-validation/observed-outcomes.tex
\begin{table*}[t]
\centering
\caption{Outcomes of the ten scripted cases, each run once under G and once under L. Groups (a) and (b) are the leak test: the worlds of a pair have the same evidence history under G until the body meets the changed fact, so G's records and every choice made before that point must agree within the pair, while L, whose rule admits the fact, follows the world. Physical results, and choices made after that point, may then differ: in (b) the interaction heads for $p_0$ in both worlds, and it shears the sheep only where the body finds it there. Groups (c) and (d) check the rules for correction, forgetting, and coverage. $\dagger$: the record contradicts the world. $p_0$, $p_1$: the position of the sheep before and after the move. Records are read through the query interface; mining and shearing are confirmed through the authoritative channel. The route cell is read from the planner's first snapshot, before the body reaches the cell, and the navigation is then cancelled. In (d) the authoritative channel places the body at each stop with the game clock stopped, and the server keeps the column loaded throughout; the four rows are checkpoints of one case, as are the two rows of the 500-tick case.}
\label{tab:boundary-full}
\scriptsize
\setlength{\tabcolsep}{8pt}
\begin{tabular}{@{}llll@{}}
\toprule
Case & Observed & G & L \\
\midrule
\multicolumn{4}{@{}l}{\textbf{(a) Hidden difference.} The site of the ore was never seen. Closure under G requires the same record and the same action at both sites.} \\
Ore sealed behind a wall, site A & record of the ore before the call; ore after it & no record; not mined & record; mined \\
Ore sealed behind a wall, site B & record of the ore before the call; ore after it & no record; not mined & record; mined \\
Control: ore seen first & record of the ore before the call; ore after it & record; mined & record; mined \\
\midrule
\multicolumn{4}{@{}l}{\textbf{(b) False belief.} The fact changed out of view after it was seen. Closure under G requires the same record and the same first choice in both worlds of a pair.} \\
Route cell left as air & cell in the planner's first snapshot & air & air \\
Route cell replaced by stone & cell in the planner's first snapshot & air$^\dagger$ & stone \\
Sheep left in place & recorded position before the call; sheep after it & $p_0$; sheared & $p_0$; sheared \\
Sheep moved six blocks & recorded position before the call; sheep after it & $p_0^\dagger$; not sheared & $p_1$; sheared \\
\midrule
\multicolumn{4}{@{}l}{\textbf{(c) Correction and forgetting.} New evidence corrects a stale record, and the retention rule drops it.} \\
Moved sheep seen again & recorded position; sheep after the call & $p_1$; sheared & $p_1$; sheared \\
Moved sheep unseen for 500 ticks & recorded position & no record & $p_1$ \\
\quad then seen again & recorded position; sheep after the call & $p_1$; sheared & $p_1$; sheared \\
\midrule
\multicolumn{4}{@{}l}{\textbf{(d) Coverage.} The body is placed far from a column, near it, and far again.} \\
Block on the surface & record at far / near / far & no / yes / yes & no / yes / no \\
Block buried below it & record at far / near / far & no / no / no & no / yes / no \\
Surface block changed to gold while far & recorded block & diamond$^\dagger$ & no record \\
\quad on return & recorded block & gold & gold \\
\bottomrule
\end{tabular}
\end{table*}

%% file: figures/sensor-recall/latex_include.tex
\begin{figure*}[!t]
\centering
\includegraphics[width=\textwidth]{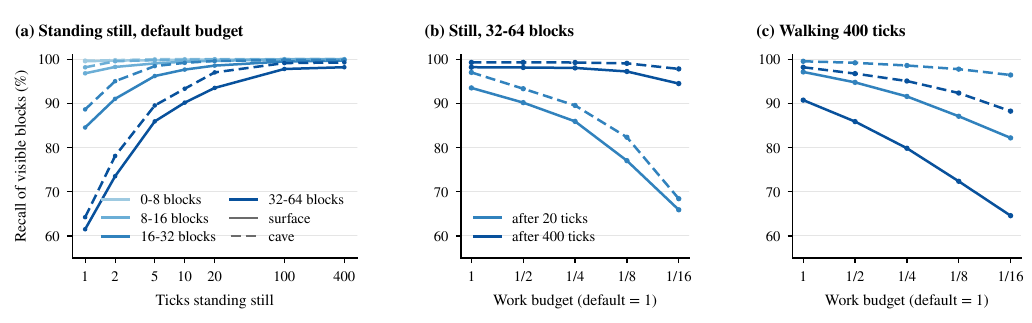}
\caption{Calibration of the G block sensor against a reference sensor with four times the work per tick on a different lattice (\V{sensor.poses} poses in \V{sensor.seeds} worlds; solid: surface poses, dashed: cave poses; poses pooled by block count). \textbf{(a)} Recall of the reference blocks while standing still, by distance from the eye, at the budget used in the experiments. \textbf{(b)} The same at 32--64 blocks after 20 and 400 ticks, as the work budget is halved four times. \textbf{(c)} Share of the reference memory held after 400 ticks of walking, for blocks first seen at 16--32 blocks (lighter) and 32--64 blocks (darker).}
\label{fig:sensor-recall}
\end{figure*}

%% file: sections/B_protocol.tex
\section{Task protocol and scoring}
\label{app:protocol}

\paragraph{Sampling and eligibility.}
Both tasks run in Minecraft 1.21.1 survival mode on easy difficulty, with an empty initial inventory and respawning enabled.
The runtime samples the authoritative state of the avatar and its surroundings at initialization, every 20 game ticks, and when an agent action ends, and sends the samples to the benchmark runner over the authority channel.
Scorers use the samples whose elapsed game time $\tau$ satisfies $0\le\tau\le T$ ($T_S=72{,}000$, $T_D=144{,}000$).

\paragraph{Survival milestone rubric.}
Table~\ref{tab:survival-rubric} lists the 22 milestones. Each milestone $j$ has a fixed point value $v_j\in[10,300]$ and a predicate evaluated on every eligible sample. Item milestones require at least one matching stack in the 36 inventory slots, either hand, or the armor slots. Portal milestones require the authoritative scanner to confirm a complete Nether portal frame, an active portal, or an End portal frame block near the actor. Location milestones require the actor to be in the Nether or inside the bounding volume of a generated fortress. Confirmations are historical and non-hierarchical: a milestone stays confirmed if the item is later consumed or lost, and confirming a higher milestone does not confirm earlier ones. $S(\tau)=\max(\{0\}\cup\{v_j:\tau_j\le\tau\})$ is a right-continuous step function, integrated exactly over $[0,T_S]$ and normalized by $300T_S$ to obtain $\mathrm{nAUC}_S$.

\begin{table*}[t]
\centering
\caption{The 22-milestone Survival rubric. ``Item held'' means the item is in the inventory, a hand, or an armor slot. Each milestone is confirmed independently from authoritative samples over $T_S=72{,}000$ simulation ticks; the score $S(\tau)$ is the largest point value among milestones confirmed by $\tau$.}
\label{tab:survival-rubric}
\small
\setlength{\tabcolsep}{3.2pt}
\begin{tabular}{@{}rlrl@{}}
\toprule
\# & Milestone & Points & Confirmation condition \\
\midrule
1 & Log or stem & 10 & Any overworld log, Nether stem, or stripped variant held (20 IDs) \\
2 & Planks & 20 & Any wood, bamboo, or Nether planks held (11 IDs) \\
3 & Crafting table & 30 & Item held \\
4 & Wooden pickaxe & 40 & Item held \\
5 & Cobblestone & 50 & Item held \\
6 & Furnace & 60 & Item held \\
7 & Stone pickaxe & 70 & Item held \\
8 & Raw iron & 90 & Item held \\
9 & Iron ingot & 100 & Item held \\
10 & Iron pickaxe & 130 & Item held \\
11 & Diamond & 150 & Item held \\
12 & Diamond pickaxe & 170 & Item held \\
13 & Obsidian & 180 & Item held \\
14 & Nether portal frame & 190 & Complete frame near the actor \\
15 & Nether portal active & 200 & Active portal near the actor \\
16 & Enter Nether & 210 & Actor in the Nether \\
17 & Enter fortress & 230 & Actor inside a Nether fortress \\
18 & Blaze rod & 250 & Item held \\
19 & Blaze powder & 260 & Item held \\
20 & Eye of ender & 280 & Item held \\
21 & End portal frame & 290 & End portal frame block near the actor \\
22 & End portal active & 300 & Active End portal near the actor \\
\bottomrule
\end{tabular}
\end{table*}

\paragraph{Discovery breadth scoring.}
Each eligible sample extracts the set of distinct canonical item IDs (\texttt{namespace:path}) with positive stack counts across the 36 main inventory slots, mainhand, offhand, and 4 armor slots. Cumulative breadth $D(\tau)$ counts the union of all canonical item IDs confirmed up to elapsed tick $\tau \le T_D$, and time-averaged breadth $\overline D = \frac{1}{T_D}\int_0^{T_D} D(\tau)\,d\tau$ is computed from the exact right-continuous step curve.

\paragraph{Agents under evaluation.}
The runtime and its agent-facing protocol (Table~\ref{tab:runtime-reads}), the regime, the worlds (Appendix~\ref{app:settings}), and the tasks, budgets, and scoring of this appendix are fixed; the model and the agent (its prompts, abilities, skills, and orchestration) may vary, and the reference agent and the tool-calling loop of Appendix~\ref{app:regime-effects} are two such agents.
Neither the model nor the agent may carry facts specific to an evaluation world, such as where ores or structures lie in it; general game knowledge, such as recipes and ore depths, is allowed.
A comparison of models holds the agent fixed (Section~\ref{sec:eval-setup}).

%% file: sections/C_regime_effects.tex
\section{World-paired effects of knowledge access}
\label{app:regime-effects}

\paragraph{Estimation and world-level consistency.}
For each task and reference configuration, we average repetitions separately under G and L within each initial world. We subtract the two world means and average the \V{design.worlds} matched-world differences with equal weights, as in Eq.~\ref{eq:paired-regime-effect}. The 95\% interval resamples the matched worlds with replacement (10{,}000 resamples, percentile method). World identity is determined by the initial snapshot. Table~\ref{tab:regime-effects-full} gives all endpoint and cumulative effects with their intervals and the signs of the individual world contrasts ($+/0/-$). Ten worlds are few for a percentile bootstrap, so the table also gives an exact sign-flip test over all \V{perm.patterns} sign patterns of the world differences. It agrees with the intervals: every effect whose interval excludes zero has $p\le\V{perm.p.sep.max}$.

\begin{figure*}[!t]
\centering
\includegraphics[width=\textwidth]{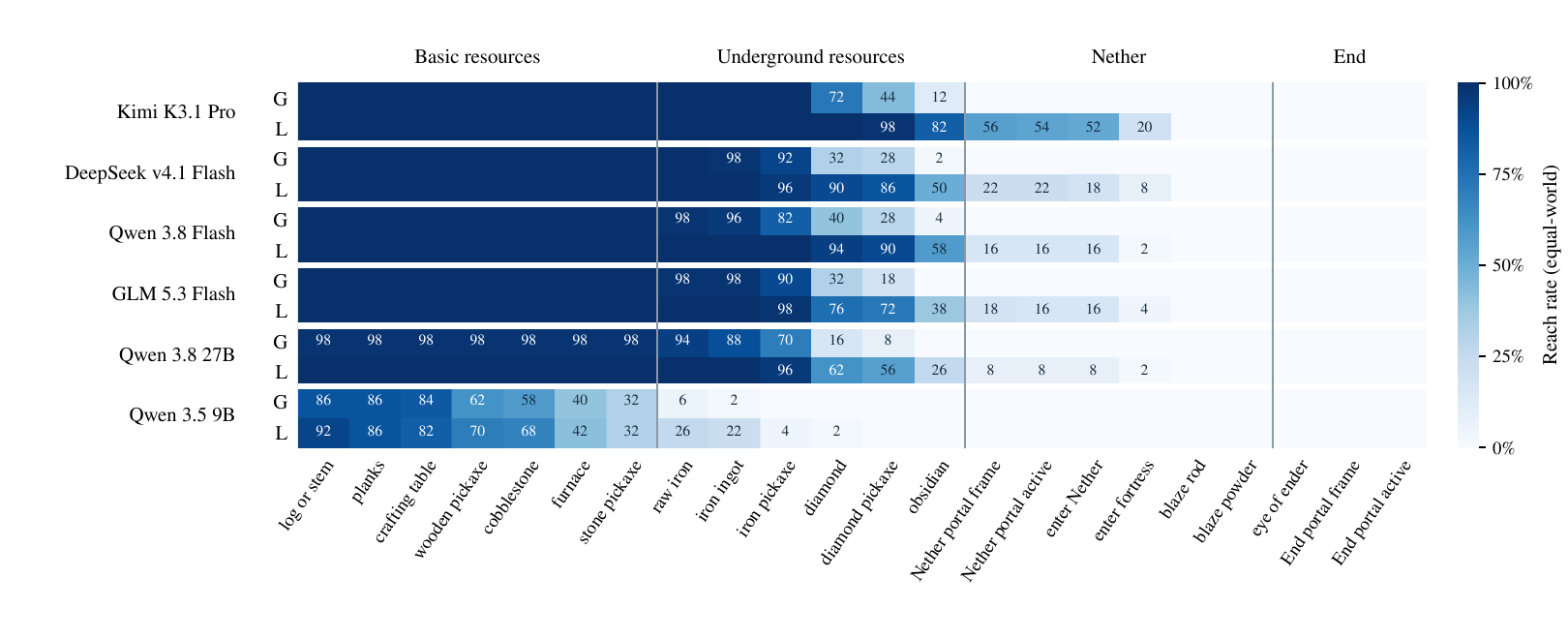}
\caption{Per-milestone Survival reach rates under perception-grounded (G) and loaded-state (L) world knowledge, shown in paired rows for each model. Color indicates the equal-world fraction of runs reaching each milestone (confirming it or a later one) within 72,000 ticks; intermediate rates are labeled as percentages. Milestones are ordered by rubric value (Table~\ref{tab:survival-rubric}).}
\label{fig:milestone-reach}
\end{figure*}

\paragraph{Milestone reach rates.}
Figure~\ref{fig:milestone-reach} reports, for every model and each of the 22 milestones, the fraction of runs that reach it within $T_S$ (repetitions averaged within each world, worlds weighted equally); Figure~\ref{fig:wall} of the main text summarizes these rates.
A run reaches a milestone when it confirms that milestone or any later one, so the reach rate of milestone $j$ is the fraction of runs with $S(T_S)\ge v_j$. Later milestones are counted because an agent may skip a prerequisite and obtain a later resource by other in-game means, such as items from village chests.
Only one of the $12\times50=600$ runs of the six models counts as reaching the diamond without holding one, and requiring the diamond itself changes the transition estimates of Table~\ref{tab:milestone-transitions} by at most \V{trans.diamond.held.maxshift} points (DeepSeek v4.1 Flash under L).

\paragraph{Milestone transitions.}
Reach rates mix the difficulty of a milestone with the time left after its prerequisite. Table~\ref{tab:milestone-transitions} therefore compares the same elapsed time: for each run that reaches the prerequisite (in the threshold sense above), the time to the next milestone is an event if reached and is censored at the end of the budget otherwise, and a Kaplan--Meier estimate gives the probability of reaching it within the window (Figure~\ref{fig:wall-crossing} for the diamond).
The estimate assumes that censoring is independent of the time to the next milestone. Censoring comes only from the budget, so runs that reach the prerequisite late are followed for less time. If late arrival predicts slower subsequent search, the estimates are optimistic~\citep{lin1999nonparametric}; if this holds more strongly under G, where prerequisites are reached later and by fewer runs, the table understates the difference between the regimes. Runs also enter at different states under the two regimes (location, equipment, explored area); the isolated stages of Appendix~\ref{app:survival-stages} remove this difference. Each bootstrap resample draws \V{design.worlds} worlds with replacement, pools their runs, and recomputes the estimate. The table lists the number of runs at risk, and intervals of $0\,[0,0]$ or $100\,[100,100]$ mark cells with no event or no failure, where the percentile bootstrap degenerates.

\input{figures/search-wall/latex_include.tex}

\input{tables/milestone-transitions/transitions.tex}

\paragraph{Doubled budget.}
The main G runs of DeepSeek v4.1 Flash have a budget of \V{budget.ticks} ticks. The agent is not told its budget (Appendix~\ref{app:runtime}), so up to $T_S$ these runs are runs with budget $T_S$, and Table~\ref{tab:budget-extension} reads the same runs at $T_S$, \V{budget.tick.mid}, and \V{budget.ticks} ticks. With the same bootstrap over worlds, the doubled budget raises their score from \V{budget.score.72k} to \V{budget.score.144k}, by \V{budget.gain} points [\V{budget.gain.low}, \V{budget.gain.high}], or \V{budget.gain.share} of the main G--L gap [\V{budget.gain.share.low}, \V{budget.gain.share.high}]. At \V{budget.ticks} ticks they still fall below the main L runs at $T_S$ in \V{budget.worlds.below} of \V{design.worlds} worlds, by \V{budget.residual} points [\V{budget.residual.low}, \V{budget.residual.high}], the remaining \V{budget.residual.share} of the gap [\V{budget.residual.share.low}, \V{budget.residual.share.high}].

\paragraph{Reduced range.}
The 2-chunk-columns row of Figure~\ref{fig:anatomy}(b) keeps L but reduces the distance within which the server delivers chunks to the client from four chunks to two; the simulation distance stays at four chunks. This retains \V{view.gap.retained} of the main G--L gap [\V{view.gap.low}, \V{view.gap.high}], and the runs end above the main G runs in \V{view.worlds.above} of \V{design.worlds} worlds.
The 32-block-sphere row instead admits only what lies within 32 blocks of the eye, the \texttt{maxDistance} of Voyager, Odyssey, and ADAM (Table~\ref{tab:code-audit}), and retains \V{r32.gap.retained} of the gap [\V{r32.gap.low}, \V{r32.gap.high}].

\paragraph{Retention window.}
The retention rows keep G and change only the number of ticks after which a block record that has not been observed again is dropped; entity records keep their window of 400 ticks.
Relative to the main G runs in the same worlds, the score shifts by \V{expiry.1.shift} points [\V{expiry.1.low}, \V{expiry.1.high}] at \V{expiry.1.label}, \V{expiry.2.shift} [\V{expiry.2.low}, \V{expiry.2.high}] at \V{expiry.2.label}, and \V{expiry.3.shift} [\V{expiry.3.low}, \V{expiry.3.high}] at \V{expiry.3.label}; the shift rises with the window, and only the interval of the shortest window excludes zero. Even that shift is about a third of the main G--L gap of \V{surv.delta.ds}.
Without block forgetting, G keeps every block it has seen until the end of the run, including blocks in chunks that L has already dropped. Main L still scores \V{budget.nf} points above it [\V{budget.nf.low}, \V{budget.nf.high}], higher in \V{budget.nf.worlds.pos} of \V{design.worlds} worlds, or \V{budget.nf.share} of the main gap [\V{budget.nf.share.low}, \V{budget.nf.share.high}]. The two configurations still differ in occlusion, in the window of entity records, and in the 64-block reach of sight, so the contrast does not isolate occlusion: the longer block memory of G works against L, and the entity window and the reach of sight work for it. The no-occlusion configuration below removes occlusion from G alone.

\paragraph{No occlusion.}
The no-occlusion row of Figure~\ref{fig:anatomy}(b) keeps G's range, update, and retention and removes only the line-of-sight test (G$^\circ$, Appendix~\ref{app:runtime}); the push, query, and tool sites all read it. Over the ten main worlds, DeepSeek v4.1 Flash scores \V{nocc.score.ds}, \V{nocc.shift.ds} [\V{nocc.shift.low.ds}, \V{nocc.shift.high.ds}] relative to the main G runs in the same worlds, or \V{nocc.gap.share.ds} of the main G--L gap of \V{surv.delta.ds}. GLM 5.3 Flash, not shown in the figure, scores \V{nocc.score.glm}, \V{nocc.shift.glm} [\V{nocc.shift.low.glm}, \V{nocc.shift.high.glm}] relative to its main G runs, or \V{nocc.gap.share.glm} of its gap of \V{surv.delta.glm}. For both models the rates of reaching the diamond, obsidian, and the Nether come close to those under L (DeepSeek v4.1 Flash: \V{nocc.diamond.ds}, \V{nocc.obsidian.ds}, \V{nocc.nether.ds} against \V{ta.L.diamond.ds}, \V{ta.L.obsidian.ds}, \V{ta.L.nether.ds}; GLM 5.3 Flash: \V{nocc.diamond.glm}, \V{nocc.obsidian.glm}, \V{nocc.nether.glm} against \V{ta.L.diamond.glm}, \V{ta.L.obsidian.glm}, \V{ta.L.nether.glm}), whereas under G \V{ta.G.diamond.ds} of the runs of each model reach the diamond and almost none reach obsidian. G$^\circ$ still differs from L in range, since its sphere does not span the full height of the world, and in retention.

\input{tables/budget-extension/budget-extension.tex}

\paragraph{Tools alone.}
Table~\ref{tab:tools-alone} gives the scores and milestone rates of the tools-alone configuration of Section~\ref{sec:anatomy}.
How to read $R$ was fixed before the runs: reads inside tools carry part of the gap if the interval of $R$ lies strictly between zero and one, and are consistent with carrying all of it if the interval excludes zero but contains one. No equivalence margin was set, so the latter does not establish full recovery.
Recovery is also uneven along the chain. The tools-alone runs of GLM 5.3 Flash reach the diamond more often than its L runs (\V{ta.A.diamond.glm} against \V{ta.L.diamond.glm}) but obsidian and the Nether less often (\V{ta.A.obsidian.glm} and \V{ta.A.nether.glm} against \V{ta.L.obsidian.glm} and \V{ta.L.nether.glm}), so its endpoint $R$ near one combines more than full recovery at the diamond with partial recovery beyond it.
In no ablation run does a turn observation or a query called by the model's program return a fact from $K_t^{\mathrm L}$; tools may still report what they found in their return values, such as the coordinates of a target.
Section~\ref{sec:anatomy} counts the tunneling calls that return a target's coordinates before breaking any block; grounding only this action's lookup of remembered targets lowers the $R$ of DeepSeek v4.1 Flash from \V{ta.A.R.ds} to \V{ta.AT.R.ds} [\V{ta.AT.R.low.ds}, \V{ta.AT.R.high.ds}].

\input{tables/tools-alone/tools-alone.tex}

\paragraph{Other harnesses.}
To test whether the regime effect depends on the reference agent, DeepSeek v4.1 Flash also runs in a second harness built on the open-source pi agent loop, which connects to the runtime through the same protocol as any third-party agent (Table~\ref{tab:harness-ladder}).
The loop makes one tool call per model step and has no program, plan, or skills. Damage and death events of the runtime are appended to the next tool result, and new damage interrupts the running action. When a request exceeds three quarters of the context window, older turns are folded into a running summary written by the same model.
The loop runs in two configurations. In the first, the tools are the abilities of the reference agent, each called directly. In the second, they are the 26 tools of the protocol and waiting, unchanged, plus four knowledge queries (block search, block lookup, region summary, and entity list) that read $K_t^\rho$ through the protocol's query method; the composites of the reference agent are not available.
Each configuration runs the \V{design.worlds} main worlds \V{design.reps} times per regime and is read at $T_S$.
In all three rows of Table~\ref{tab:harness-ladder}, the reference agent included, the effect is positive in every world (exact sign-flip $p=\V{harness.signflip.p}$).
Absolute scores are not comparable across rows: the harnesses differ in prompts, skills, and tools.

\begin{table*}[!t]
\centering
\caption{Survival regime effect of DeepSeek v4.1 Flash in two harnesses. The first row is the main runs. Scores are means over worlds, L$-$G is the world-paired effect with a 95\% bootstrap interval over worlds, and a run reaches a milestone when it confirms it or a later one within $T_S$; worlds are weighted equally.}
\label{tab:harness-ladder}
\small
\setlength{\tabcolsep}{4pt}
\begin{tabular}{@{}llrrrcccc@{}}
\toprule
& & & & & & \multicolumn{3}{c}{Runs reaching, G / L (\%)} \\
\cmidrule(l){7-9}
Harness & Tools & G & L & L$-$G [95\% interval] & Worlds $+/0/-$ & Diamond & Obsidian & Nether \\
\midrule
Reference agent & Abilities, in programs & \V{harness.ref.G} & \V{harness.ref.L} & \V{harness.ref.delta} [\V{harness.ref.low}, \V{harness.ref.high}] & \V{harness.ref.worlds.pos}/\V{harness.ref.worlds.tie}/\V{harness.ref.worlds.neg} & \V{harness.ref.diamond.G} / \V{harness.ref.diamond.L} & \V{harness.ref.obsidian.G} / \V{harness.ref.obsidian.L} & \V{harness.ref.nether.G} / \V{harness.ref.nether.L} \\
Tool-calling loop & Abilities & \V{harness.abil.G} & \V{harness.abil.L} & \V{harness.abil.delta} [\V{harness.abil.low}, \V{harness.abil.high}] & \V{harness.abil.worlds.pos}/\V{harness.abil.worlds.tie}/\V{harness.abil.worlds.neg} & \V{harness.abil.diamond.G} / \V{harness.abil.diamond.L} & \V{harness.abil.obsidian.G} / \V{harness.abil.obsidian.L} & \V{harness.abil.nether.G} / \V{harness.abil.nether.L} \\
Tool-calling loop & Protocol tools and queries & \V{harness.proto.G} & \V{harness.proto.L} & \V{harness.proto.delta} [\V{harness.proto.low}, \V{harness.proto.high}] & \V{harness.proto.worlds.pos}/\V{harness.proto.worlds.tie}/\V{harness.proto.worlds.neg} & \V{harness.proto.diamond.G} / \V{harness.proto.diamond.L} & \V{harness.proto.obsidian.G} / \V{harness.proto.obsidian.L} & \V{harness.proto.nether.G} / \V{harness.proto.nether.L} \\
\bottomrule
\end{tabular}
\end{table*}

\input{tables/regime-effects/paired-effects.tex}

%% file: figures/search-wall/latex_include.tex
\begin{figure}[!htb]
\centering
\includegraphics[width=\columnwidth]{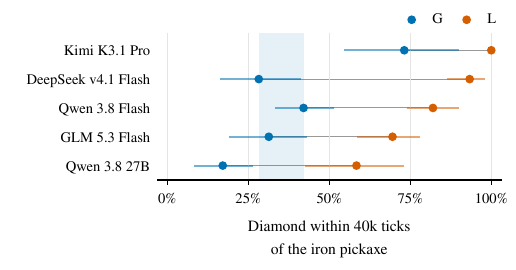}
\caption{Crossing the wall. Kaplan--Meier probability of reaching the diamond within \V{trans.window} ticks of reaching the iron pickaxe (reach: the milestone or a later one is confirmed), with 95\% intervals, for models with at least ten such runs per regime. The band spans the Flash models under G.}
\label{fig:wall-crossing}
\end{figure}

%% file: tables/milestone-transitions/transitions.tex
\begin{table*}[!t]
\centering
\caption{Percentage of runs that reach a milestone (confirm it or a later one) within 40{,}000 ticks of its prerequisite (Kaplan--Meier, censored at the end of the budget), with 95\% bootstrap intervals over worlds and, in parentheses, the number of runs at risk. Runs that never reach the prerequisite are not at risk; a dash marks a model with no such run. Intervals of 0 [0, 0] and 100 [100, 100] occur where no world has an event or a failure and the percentile bootstrap degenerates.}
\label{tab:milestone-transitions}
\small
\begin{tabular}{@{}lcccc@{}}
\toprule
& \multicolumn{2}{c}{Iron pickaxe $\to$ diamond} & \multicolumn{2}{c}{Diamond pickaxe $\to$ obsidian} \\
\cmidrule(lr){2-3}\cmidrule(l){4-5}
Model & G & L & G & L \\
\midrule
Kimi K3.1 Pro & 73 [55, 90] (50) & 100 [100, 100] (50) & 21 [0, 52] (22) & 85 [72, 95] (49) \\
DeepSeek v4.1 Flash & 28 [16, 41] (46) & 93 [86, 98] (48) & 7 [0, 21] (14) & 55 [41, 69] (43) \\
Qwen 3.8 Flash & 42 [33, 52] (41) & 82 [74, 90] (50) & 16 [0, 42] (14) & 65 [44, 82] (45) \\
GLM 5.3 Flash & 31 [19, 43] (45) & 70 [59, 78] (49) & 0 [0, 0] (9) & 56 [33, 78] (36) \\
Qwen 3.8 27B & 17 [8, 26] (35) & 58 [43, 73] (48) & 0 [0, 0] (4) & 45 [28, 64] (28) \\
Qwen 3.5 9B & -- & 50 [0, 100] (2) & -- & -- \\
\bottomrule
\end{tabular}
\end{table*}

%% file: tables/budget-extension/budget-extension.tex
\begin{table*}[!t]
\centering
\caption{Sensitivity of the Survival gap of DeepSeek v4.1 Flash. The main G runs extend the budget to 144,000 ticks; the main results read them at 72,000 ticks, and this table also at 108,000 and 144,000. The block-expiry G runs drop a block record when the block has not been observed again for the stated window (80 ticks, 2,000 ticks, no forgetting); the two-chunk L runs reduce the delivery boundary from four chunks to two. Apart from the readings of the main G runs, each row is a separate set of runs on the worlds and configuration of the main runs apart from the changed setting. Scores and rates weight worlds equally.}
\label{tab:budget-extension}
\small
\begin{tabular}{@{}lrrrrr@{}}
\toprule
& & & \multicolumn{3}{c}{Runs reaching (\%)} \\
\cmidrule(l){4-6}
Runs & Tick & Score & Diamond & Obsidian & Nether \\
\midrule
G, main runs & 72,000 & 139.6 & 32 & 2 & 0 \\
G, main runs & 108,000 & 143.6 & 40 & 2 & 0 \\
G, main runs & 144,000 & 147.0 & 46 & 2 & 0 \\
G, retention 80 ticks & 72,000 & 126.0 & 24 & 0 & 0 \\
G, retention 2,000 ticks & 72,000 & 137.8 & 36 & 2 & 2 \\
G, retention no forgetting & 72,000 & 140.6 & 38 & 10 & 2 \\
L, main runs & 72,000 & 176.8 & 90 & 50 & 18 \\
L, two-chunk range & 72,000 & 178.8 & 92 & 52 & 20 \\
\bottomrule
\end{tabular}
\end{table*}

%% file: tables/tools-alone/tools-alone.tex
\begin{table*}[!t]
\centering
\caption{Survival tools-alone ablation. Pushed observations and the model's own queries read $K_t^{\mathrm G}$ and reads inside tools read $K_t^{\mathrm L}$; the tunnel-lookup variant also grounds the tunneling action's lookup of remembered targets. $R$ is the share of the G--L score gap recovered, with a 95\% bootstrap interval over worlds. Scores and rates weight worlds equally.}
\label{tab:tools-alone}
\small
\begin{tabular}{@{}llrrrrr@{}}
\toprule
& & & & \multicolumn{3}{c}{Runs reaching (\%)} \\
\cmidrule(l){5-7}
Model & Condition & Score & $R$ [95\% interval] & Diamond & Obsidian & Nether \\
\midrule
DeepSeek v4.1 Flash & G & 139.6 & -- & 32 & 2 & 0 \\
 & Tools alone & 165.6 & 0.70 [0.48, 0.92] & 80 & 28 & 6 \\
 & Tools alone, tunnel lookup in G & 157.4 & 0.48 [0.24, 0.74] & 62 & 14 & 8 \\
 & L & 176.8 & -- & 90 & 50 & 18 \\
\midrule
GLM 5.3 Flash & G & 136.4 & -- & 32 & 0 & 0 \\
 & Tools alone & 168.0 & 0.98 [0.81, 1.21] & 88 & 26 & 4 \\
 & L & 168.6 & -- & 76 & 38 & 16 \\
\bottomrule
\end{tabular}
\end{table*}

%% file: tables/regime-effects/paired-effects.tex
\begin{table*}[t]
\centering\small
\caption{World-paired effects for final and cumulative performance under perception-grounded (G) and loaded-state (L) world knowledge. G and L entries first average repetitions within each initial world, then average matched worlds equally. The 95\% interval is a percentile bootstrap over worlds (10{,}000 resamples); Worlds $+/0/-$ counts positive, zero, and negative within-world L-minus-G differences. $p$ is an exact two-sided sign-flip test of a zero mean over all sign patterns of the world differences, which assumes only that, without an effect, each world difference is symmetric about zero. Survival points follow the predefined progression rubric; $\mathrm{nAUC}_S$ is dimensionless; both Discovery metrics have units of confirmed item types.}
\label{tab:regime-effects-full}
\setlength{\tabcolsep}{4pt}
\begin{tabular}{@{}llrrrccr@{}}
\toprule
Configuration & Metric & G & L & $\Delta_{\mathrm{L}-\mathrm{G}}$ & 95\% interval & Worlds $+/0/-$ & $p$\\
\midrule
\multicolumn{8}{l}{\textbf{Survival}}\\
Kimi K3.1 Pro & $S(T_S)$ & 154.4 & 198.0 & $+43.6$ & $[+36.4, +50.8]$ & 10/0/0 & 0.002\\
 & $\mathrm{nAUC}_S$ & 0.431 & 0.538 & $+0.107$ & $[+0.084, +0.130]$ & 10/0/0 & 0.002\\
DeepSeek v4.1 Flash & $S(T_S)$ & 139.6 & 176.8 & $+37.2$ & $[+30.8, +44.2]$ & 10/0/0 & 0.002\\
 & $\mathrm{nAUC}_S$ & 0.398 & 0.509 & $+0.111$ & $[+0.090, +0.134]$ & 10/0/0 & 0.002\\
Qwen 3.8 Flash & $S(T_S)$ & 137.8 & 177.8 & $+40.0$ & $[+33.0, +46.6]$ & 10/0/0 & 0.002\\
 & $\mathrm{nAUC}_S$ & 0.392 & 0.509 & $+0.116$ & $[+0.098, +0.133]$ & 10/0/0 & 0.002\\
GLM 5.3 Flash & $S(T_S)$ & 136.4 & 168.6 & $+32.2$ & $[+25.4, +40.0]$ & 10/0/0 & 0.002\\
 & $\mathrm{nAUC}_S$ & 0.382 & 0.483 & $+0.101$ & $[+0.066, +0.144]$ & 10/0/0 & 0.002\\
Qwen 3.8 27B & $S(T_S)$ & 122.0 & 157.8 & $+35.8$ & $[+24.6, +49.6]$ & 10/0/0 & 0.002\\
 & $\mathrm{nAUC}_S$ & 0.351 & 0.439 & $+0.088$ & $[+0.045, +0.146]$ & 10/0/0 & 0.002\\
Qwen 3.5 9B & $S(T_S)$ & 46.2 & 56.2 & $+10.0$ & $[-0.4, +23.2]$ & 6/1/3 & 0.16\\
 & $\mathrm{nAUC}_S$ & 0.138 & 0.152 & $+0.014$ & $[-0.011, +0.041]$ & 5/0/5 & 0.34\\
\addlinespace[4pt]
\multicolumn{8}{l}{\textbf{Discovery}}\\
Kimi K3.1 Pro & $D(T_D)$ & 136.8 & 171.9 & $+35.1$ & $[+29.9, +40.5]$ & 10/0/0 & 0.002\\
 & $\overline D$ & 82.0 & 105.3 & $+23.3$ & $[+8.6, +38.0]$ & 7/0/3 & 0.021\\
DeepSeek v4.1 Flash & $D(T_D)$ & 118.7 & 132.6 & $+13.9$ & $[+5.7, +23.3]$ & 9/0/1 & 0.014\\
 & $\overline D$ & 81.9 & 90.8 & $+8.9$ & $[+3.3, +14.6]$ & 7/0/3 & 0.025\\
Qwen 3.8 Flash & $D(T_D)$ & 93.8 & 113.6 & $+19.8$ & $[+14.4, +25.8]$ & 10/0/0 & 0.002\\
 & $\overline D$ & 64.1 & 76.8 & $+12.7$ & $[+10.0, +15.5]$ & 10/0/0 & 0.002\\
GLM 5.3 Flash & $D(T_D)$ & 69.0 & 83.5 & $+14.5$ & $[+8.4, +20.8]$ & 10/0/0 & 0.002\\
 & $\overline D$ & 45.8 & 58.0 & $+12.2$ & $[+7.8, +16.4]$ & 9/0/1 & 0.004\\
Qwen 3.8 27B & $D(T_D)$ & 54.5 & 74.5 & $+20.0$ & $[+16.1, +24.2]$ & 10/0/0 & 0.002\\
 & $\overline D$ & 41.1 & 52.4 & $+11.3$ & $[+8.0, +14.8]$ & 10/0/0 & 0.002\\
Qwen 3.5 9B & $D(T_D)$ & 19.4 & 18.8 & $-0.6$ & $[-4.5, +4.2]$ & 4/0/6 & 0.78\\
 & $\overline D$ & 14.4 & 13.0 & $-1.4$ & $[-4.6, +2.5]$ & 3/0/7 & 0.48\\
\bottomrule
\end{tabular}
\end{table*}

%% file: sections/D_isolated_stages.tex
\section{Isolated Survival stages}
\label{app:survival-stages}

\input{tables/stage-diagnostics/stage-specifications.tex}

\paragraph{Task protocol.}
Every world uses the same seed. Each isolated stage starts from a prescribed location, with the prescribed inventory and equipment (Table~\ref{tab:stage-diagnostics-specs}) and with empty dialogue history and dynamic world knowledge. Instructions omit target coordinates. Stage 8 requires portal construction, activation, and entry into the Nether. Stage 10 instructs the agent to obtain blaze powder and supplies a bow and sixty-four arrows in addition to the combat kit. Its initial score is 230; a first confirmed blaze rod yields 250, and any confirmed blaze powder yields 260 and completes the stage. Stage 11 supplies twenty-four blaze powder and twenty-four ender pearls and requires End-portal activation.

Stage scores follow the rubric values, except that Stages 1, 3, and 7 credit quantities toward their targets: $10/3$ points per log up to three, $10/11$ per cobblestone up to eleven, and one point per obsidian up to ten; Stage 4 credits the furnace and the stone pickaxe 10 points each, in either order. Scoring-endpoint confirmation, first confirmed death, or budget exhaustion ends an execution. Death takes precedence over new progress in the same sample, and initialized supplies do not confirm intermediate milestones. These conditions differ from full-horizon Survival, which allows respawning. Running a stage in isolation reduces dependence on preceding milestones; it also changes the starting resources and goal context, so the stages do not form a natural full-task trajectory. Table~\ref{tab:stage-diagnostics-outcomes} gives the outcome counts of every stage.

\input{tables/stage-diagnostics/stage-outcomes.tex}

\paragraph{Second agent.}
GLM 5.3 Flash runs the search stages and the combat stage in the same scenes, with the same protocols and budgets (Table~\ref{tab:stage-outcomes-second-model}).
Summed over the search stages, it succeeds less often than DeepSeek v4.1 Flash under both regimes, and its failures under G divide into \V{stage.glm.search.G.timeout} timeouts and \V{stage.glm.search.G.death} deaths. It was not run on the non-search Stages 1--4 and 8.

\input{tables/stage-diagnostics/stage-outcomes-second-model.tex}

\input{tables/stage-failure-causes/death-causes.tex}

\paragraph{Human reference in the diamond stage.}
Thirty human operators each run Stage 6 once under G, with the start, supplies, instruction, and budget of the model executions (Table~\ref{tab:stage-diagnostics-specs}).
An operator acts only by calling the reference agent's abilities through the same protocol.
Operators are given a web interface in which they call tools to obtain a 3D structural rendering of the world knowledge $K_t^{\mathrm G}$ instead of structured data, and the coordinates are visible in this rendering; all other operations, including those that return rendered frames, are the same as the agent's.
Operators were unpaid adult volunteers, recruited through personal contacts, who knew the basic rules of Minecraft. Before their run, each read a written guide to the web interface; none had a practice run. No ethics review was sought: operators only played a game task through a web interface, no personal data were collected, and the recorded outcomes cannot be linked to individuals. All operators consented to the use of their results.
Operators complete the stage in \V{stage.6.human.success} of \V{stage.6.human.n} executions; the other \V{stage.6.human.timeout} time out and none die.
Two-sided Fisher exact tests compare these counts with those of DeepSeek v4.1 Flash under G (\V{stage.6.G.success} of \V{stage.6.G.n}, $p=\V{stage.6.human.p.G}$) and under L (\V{stage.6.L.success} of \V{stage.6.L.n}, $p=\V{stage.6.human.p.L}$).

\paragraph{Causes of death.}
Table~\ref{tab:stage-death-causes} gives the causes of death.
In Stage 10, \V{stage.10.G.success} G and \V{stage.10.L.success} L executions confirm blaze powder; the remaining \V{stage.10.G.death} G and \V{stage.10.L.death} L executions end in death, with no timeouts. L's mean final score is \V{stage.10.L.score}, compared with \V{stage.10.G.score} in G. Fireball damage, which in this scene comes from blazes, first occurs after a median of \V{combat.firstfireball.L} ticks under L and \V{combat.firstfireball.G} under G (over the \V{combat.fireballruns.L} and \V{combat.fireballruns.G} executions that take any). L thus brings the agent to its targets sooner, yet completion stays low: once targets are known, the bottleneck moves from finding to fighting through discrete, program-level calls. GLM 5.3 Flash completes \V{stage.glm.10.G.success} and \V{stage.glm.10.L.success} of \V{stage.glm.10.G.n} executions under G and L, and all of its other executions end in death.

\paragraph{Stage curves.}
Figure~\ref{fig:survival-stages} plots the mean stage score over time. Successful executions and executions ending in a first death keep their final scores until the end of the stage budget and remain in the denominator. Each stage occupies an equal-width segment on its own time scale, and only tails that are constant under both regimes are trimmed.

\input{figures/survival-stages/latex_include.tex}

\begin{figure*}[!t]
\centering
\begin{minipage}[t]{0.48\textwidth}
\vspace{0pt}
\centering
\includegraphics[height=0.85\linewidth]{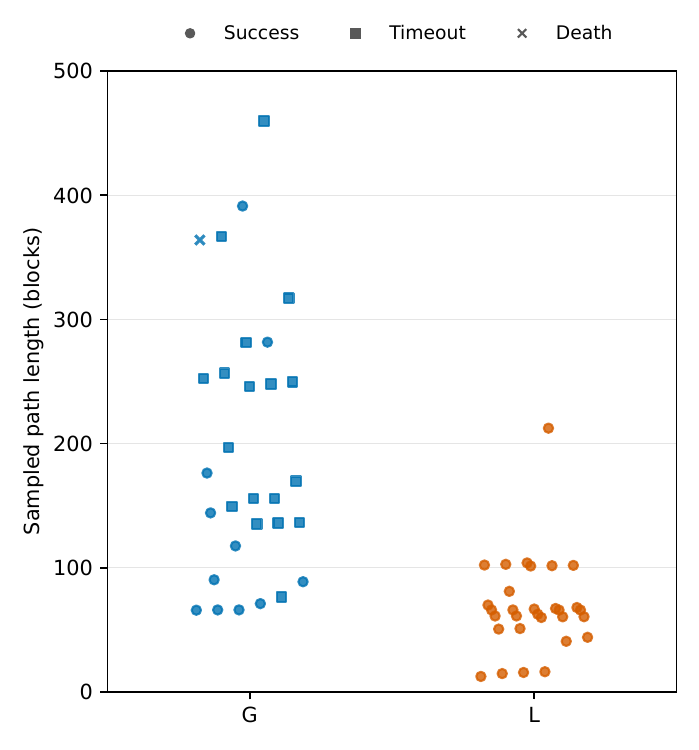}
\caption{Sampled path lengths of all executions in the diamond-pickaxe stage (Stage 6), by regime and outcome.}
\label{fig:stage6-paths}
\end{minipage}\hfill
\begin{minipage}[t]{0.48\textwidth}
\vspace{0pt}
\centering
\includegraphics[height=0.85\linewidth]{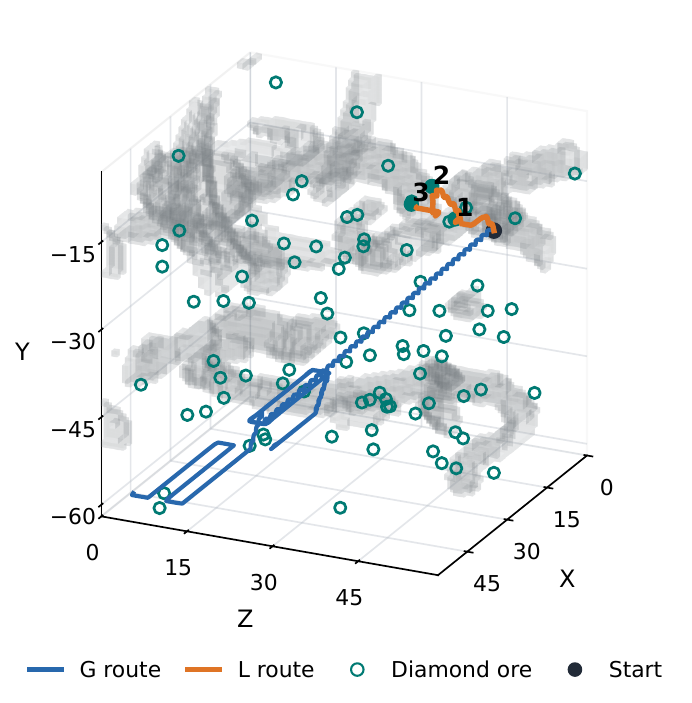}
\caption{Illustrative routes under G (blue) and L (orange) from a shared start (dark circle), one execution per regime. Hollow circles mark diamond ore, and numbers mark the sites mined under L in order. Gray surfaces show the initial cave boundary.}
\label{fig:diamond-route-case}
\end{minipage}
\end{figure*}

\paragraph{Search paths in the diamond stage.}
Figure~\ref{fig:stage6-paths} uses all Stage 6 executions, which share world, start, supplies, and instruction across regimes. Body positions are read from the authoritative observation log; the sampled path length sums 3D distances between consecutive samples within a dimension while the actor is alive. The median G execution travels \V{stage6.path.G} blocks (interquartile range \V{stage6.pathq1.G}--\V{stage6.pathq3.G}) and reaches $Y{=}$\V{stage6.miny.G}; the median L execution travels \V{stage6.path.L} blocks (\V{stage6.pathq1.L}--\V{stage6.pathq3.L}) and reaches $Y{=}$\V{stage6.miny.L}.

\paragraph{Illustrative routes.}
Figure~\ref{fig:diamond-route-case} shows a paired G execution and L execution. Under G, the agent travels about \V{routecase.G.dist} blocks over its full budget without obtaining a diamond; under L, it mines three nearby ore sites and completes the pickaxe in \V{routecase.L.ticks} ticks with about \V{routecase.L.dist} blocks of travel. The figure illustrates the difference between searching and target-guided collection.

%% file: tables/stage-diagnostics/stage-specifications.tex
\begin{table*}[t]
\centering
\caption{Protocol and environment specifications for the eleven isolated Survival stages. Within each stage, G/L share initial snapshot, coordinates, and supplies. \emph{Start}: where the agent is placed; Stages 1--4 use the world spawn. $^\dagger$Mining kit contains 8 oak logs, 1 crafting table, 1 furnace, 8 coal, 32 cobblestone, 32 torches, 16 cooked beef, and 128 dirt. $^\ddagger$Combat kit contains 1 iron sword, 1 iron pickaxe, 8 oak logs, 1 crafting table, 64 cobblestone, 32 torches, 32 cooked beef, and 128 dirt, with an equipped iron chestplate and shield. Stage 10 additionally supplies a bow and 64 arrows; its endpoint requires any confirmed blaze powder.}
\label{tab:stage-diagnostics-specs}
\small
\setlength{\tabcolsep}{4.5pt}
\begin{tabularx}{\textwidth}{@{}clccl>{\raggedright\arraybackslash}X>{\raggedright\arraybackslash}p{4.2cm}@{}}
\toprule
Stage & Milestone Target & Score & Budget & Start & Task Instruction Prompt & Prescribed Supplies Highlights \\
\midrule
1 & Logs & $0 \to 10$ & 2,400 & Surface & \textit{``Obtain 3 logs.''} & None (bare hand) \\
2 & Wooden Pickaxe & $10 \to 40$ & 2,400 & Surface & \textit{``Obtain a wooden pickaxe.''} & 3 oak logs \\
3 & Cobblestone & $40 \to 50$ & 3,600 & Surface & \textit{``Obtain 11 cobblestone.''} & 1 wooden pickaxe \\
4 & Furnace + Stone Pickaxe & $50 \to 70$ & 2,400 & Surface & \textit{``Obtain a furnace and a stone pickaxe, in either order.''} & 11 cobblestone, 1 crafting table, 2 sticks \\
5 & Iron Pickaxe & $70 \to 130$ & 12,000 & Cave & \textit{``Obtain an iron pickaxe.''} & 1 stone pickaxe, mining kit$^\dagger$ \\
6 & Diamond Pickaxe & $130 \to 170$ & 18,000 & Cave & \textit{``Obtain a diamond pickaxe.''} & 1 iron pickaxe, mining kit$^\dagger$ \\
7 & Obsidian & $170 \to 180$ & 12,000 & Cave & \textit{``Obtain 10 obsidian.''} & 1 diamond pickaxe, 1 water bucket, 3 iron ingots, mining kit$^\dagger$ \\
8 & Enter Nether & $180 \to 210$ & 6,000 & Cave & \textit{``Build and activate a Nether portal, then enter the Nether.''} & 10 obsidian, 1 diamond pickaxe, 1 flint and steel, basic blocks, food \\
9 & Nether Fortress & $210 \to 230$ & 18,000 & Nether & \textit{``Find and enter a Nether fortress.''} & Combat kit$^\ddagger$, 1 flint and steel \\
10 & Blaze Powder & $230 \to 260$ & 18,000 & Nether & \textit{``Obtain blaze powder.''} & Combat kit$^\ddagger$, 1 bow, 64 arrows \\
11 & End Portal & $260 \to 300$ & 24,000 & Surface & \textit{``Activate an End portal. Entering it is not required.''} & Combat kit$^\ddagger$, 24 blaze powder, 24 ender pearls, 1 water bucket \\
\bottomrule
\end{tabularx}
\end{table*}

%% file: tables/stage-diagnostics/stage-outcomes.tex
\begin{table*}[t]
\centering
\caption{Outcomes of the isolated Survival stages for DeepSeek v4.1 Flash, with thirty executions per regime. Budget is measured in simulation ticks; score is mean final confirmed stage score. S / D / T gives scoring-endpoint, first-death, and timeout counts. Stage 8 requires Nether entry; Stage 10 supplies a bow and arrows and requires any blaze powder.}
\label{tab:stage-diagnostics-outcomes}
\small
\setlength{\tabcolsep}{5pt}
\begin{tabular}{@{}clcrrcrc@{}}
\toprule
& & & & \multicolumn{2}{c}{\textbf{G}} & \multicolumn{2}{c}{\textbf{L}} \\
\cmidrule(lr){5-6} \cmidrule(l){7-8}
Stage & Milestone target & Score range & Budget & Score & S / D / T & Score & S / D / T \\
\midrule
1 & Logs & $0 \to 10$ & 2,400 & 10.0 & 30 / 0 / 0 & 10.0 & 30 / 0 / 0 \\
2 & Wooden Pickaxe & $10 \to 40$ & 2,400 & 40.0 & 30 / 0 / 0 & 40.0 & 30 / 0 / 0 \\
3 & Cobblestone & $40 \to 50$ & 3,600 & 50.0 & 30 / 0 / 0 & 50.0 & 30 / 0 / 0 \\
4 & Furnace + Stone Pickaxe & $50 \to 70$ & 2,400 & 70.0 & 30 / 0 / 0 & 70.0 & 30 / 0 / 0 \\
5 & Iron Pickaxe & $70 \to 130$ & 12,000 & 104.0 & 15 / 9 / 6 & 130.0 & 30 / 0 / 0 \\
6 & Diamond Pickaxe & $130 \to 170$ & 18,000 & 148.7 & 11 / 1 / 18 & 170.0 & 30 / 0 / 0 \\
7 & Obsidian & $170 \to 180$ & 12,000 & 173.2 & 8 / 3 / 19 & 177.5 & 19 / 8 / 3 \\
8 & Enter Nether & $180 \to 210$ & 6,000 & 208.7 & 28 / 2 / 0 & 209.3 & 29 / 1 / 0 \\
9 & Nether Fortress & $210 \to 230$ & 18,000 & 214.7 & 7 / 19 / 4 & 225.3 & 23 / 7 / 0 \\
10 & Blaze Powder & $230 \to 260$ & 18,000 & 232.0 & 2 / 28 / 0 & 238.3 & 7 / 23 / 0 \\
11 & End Portal & $260 \to 300$ & 24,000 & 280.0 & 0 / 1 / 29 & 297.0 & 23 / 6 / 1 \\
\bottomrule
\end{tabular}
\end{table*}

%% file: tables/stage-diagnostics/stage-outcomes-second-model.tex
\begin{table*}[t]
\centering
\caption{Outcomes of GLM 5.3 Flash in the isolated search stages and the combat stage, with 30 executions per regime in the scenes, protocols, and budgets of Table~\ref{tab:stage-diagnostics-specs}. Columns as in Table~\ref{tab:stage-diagnostics-outcomes}.}
\label{tab:stage-outcomes-second-model}
\small
\setlength{\tabcolsep}{5pt}
\begin{tabular}{@{}clrcrc@{}}
\toprule
& & \multicolumn{2}{c}{\textbf{G}} & \multicolumn{2}{c}{\textbf{L}} \\
\cmidrule(lr){3-4} \cmidrule(l){5-6}
Stage & Milestone target & Score & S / D / T & Score & S / D / T \\
\midrule
5 & Iron Pickaxe & 108.7 & 17 / 9 / 4 & 124.7 & 27 / 3 / 0 \\
6 & Diamond Pickaxe & 143.3 & 9 / 1 / 20 & 168.0 & 28 / 1 / 1 \\
7 & Obsidian & 171.6 & 2 / 16 / 12 & 178.0 & 21 / 9 / 0 \\
9 & Nether Fortress & 210.7 & 1 / 28 / 1 & 222.7 & 19 / 11 / 0 \\
10 & Blaze Powder & 232.0 & 2 / 28 / 0 & 233.0 & 3 / 27 / 0 \\
11 & End Portal & 280.0 & 0 / 2 / 28 & 289.3 & 10 / 7 / 13 \\
\bottomrule
\end{tabular}
\end{table*}

%% file: tables/stage-failure-causes/death-causes.tex
\begin{table*}[!t]
\centering
\caption{Cause of death of DeepSeek v4.1 Flash in the isolated Survival stages (stages with at least one death). The cause is the damage type of the event that reduced health to zero, from the harness damage log; \emph{Mob} combines melee, projectile, fireball, explosion, and wither damage; \emph{Fire} combines burning and standing in fire.}
\label{tab:stage-death-causes}
\small
\begin{tabular}{@{}crrrrrrrrrr@{}}
\toprule
& \multicolumn{5}{c}{\textbf{G}} & \multicolumn{5}{c}{\textbf{L}} \\
\cmidrule(lr){2-6}\cmidrule(l){7-11}
Stage & \textbf{Deaths} & Lava & Fire & Fall & Mob & \textbf{Deaths} & Lava & Fire & Fall & Mob \\
\midrule
5 & \textbf{9} & 0 & 0 & 0 & 9 & \textbf{0} & 0 & 0 & 0 & 0 \\
6 & \textbf{1} & 0 & 0 & 0 & 1 & \textbf{0} & 0 & 0 & 0 & 0 \\
7 & \textbf{3} & 2 & 1 & 0 & 0 & \textbf{8} & 2 & 0 & 0 & 6 \\
8 & \textbf{2} & 0 & 0 & 0 & 2 & \textbf{1} & 0 & 0 & 0 & 1 \\
9 & \textbf{19} & 5 & 2 & 4 & 8 & \textbf{7} & 1 & 1 & 0 & 5 \\
10 & \textbf{28} & 3 & 4 & 1 & 20 & \textbf{23} & 2 & 3 & 1 & 17 \\
11 & \textbf{1} & 0 & 0 & 0 & 1 & \textbf{6} & 3 & 0 & 0 & 3 \\
\bottomrule
\end{tabular}
\end{table*}

%% file: figures/survival-stages/latex_include.tex
\begin{figure*}[!t]
\centering
\includegraphics[width=0.85\textwidth]{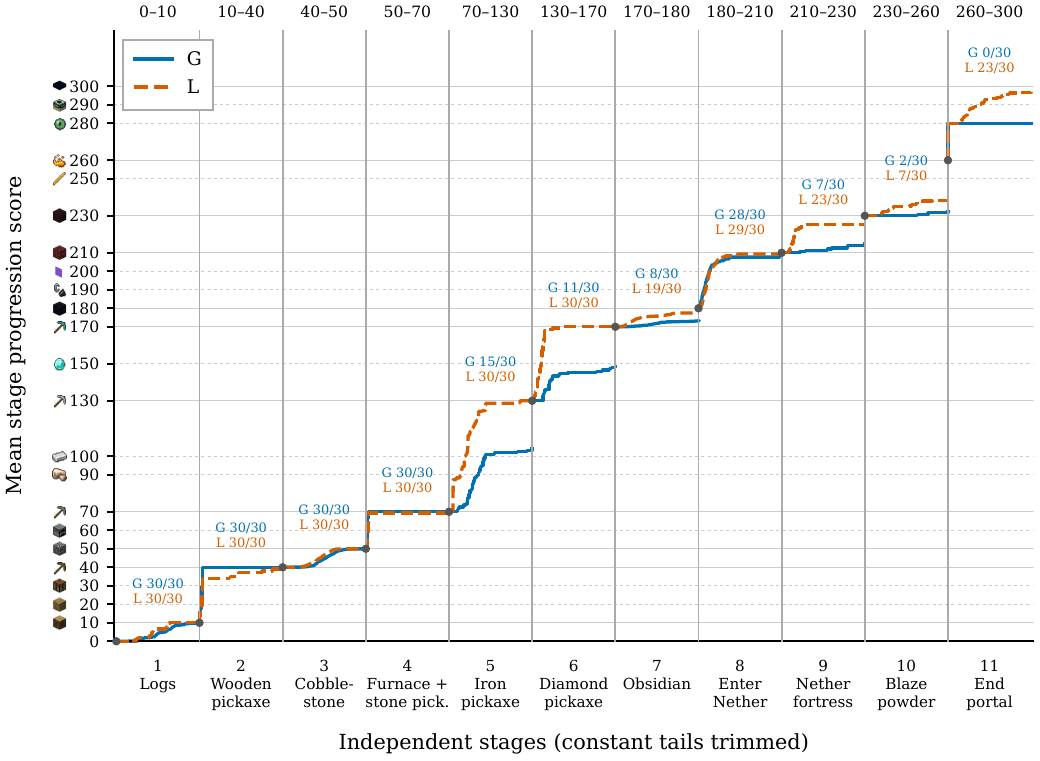}
\caption{Stage-wise Survival progression of DeepSeek v4.1 Flash under G (solid blue) and L (dashed orange). Curves are mean stage scores over time; colored counts give the executions that reach the stage endpoint. Each stage spans its prescribed start and goal scores, which differ from full-Survival scores; stages are independent runs with different time scales. Figure~\ref{fig:anatomy}(a) gives the outcome counts of the same executions.}
\label{fig:survival-stages}
\end{figure*}

%% file: sections/E_discovery.tex
\section{Discovery composition}
\label{app:discovery-diagnostics}

\paragraph{Dependency categories.}
We grade every item by how strongly its acquisition depends on world knowledge: surface items can be found by looking around, mobs move and can be occluded, and subsurface resources are hidden by terrain.
Each item receives a dependency level of 0, 1, or 2, computed over its acquisition routes in the vanilla 1.21.1 data pack (recipes, tags, loot tables, and world generation).
Hand-written lists cover what the data pack does not encode: natural blocks placed by hard-coded features (e.g., glowstone, bamboo), deep-layer blocks (e.g., deepslate, obsidian), and about forty interactions (e.g., filling buckets, milking, concrete hardening, copper oxidation, log stripping).
\emph{Regular routes} are block drops, mob drops, recipes, and interactions.
A block source has level~2 if the block is an ore, a deep-layer block, or native to the Nether or End, and level~0 otherwise; a mob drop has level~1, or level~2 for Nether and End mobs.
Golems, the Ender Dragon, and the Wither are excluded.
Each ingredient slot takes the minimum level over its alternatives (e.g., the members of a tag); a recipe or interaction takes the maximum over its slots and required station, and an item takes the minimum over its routes, iterated to a fixed point. Shaped recipes wider or taller than two cells and shapeless recipes with more than four ingredients require a crafting table; processing uses the station of its recipe type (furnace, blast furnace, smoker, campfire, stonecutter, or smithing table). Quantities and fuel are ignored, mining tools are not ingredients, and non-tool probability and context conditions count as satisfiable. Regular harvesting uses bare hands or shears, without Silk Touch or Looting.
\emph{Fallback routes} are loot tables other than block and mob drops (chests, archaeology, fishing, and others), at level~1 wherever the structure lies; mob drops gated by a random chance or tied to rare structures and events (e.g., witches, ravagers, villagers), at the mob's level; and block drops that require Silk Touch, at the block's level.
They are admitted in rounds, only for items that are still unreached: each round fixes the levels of all reachable items, so a fallback never lowers them, and serves items without any regular route first.
Special recipes without a fixed result are not modeled; suspicious stew therefore counts as loot.
Level~2 items are \emph{subsurface-dependent} (\V{disc.taxonomy.subsurface} of the \V{disc.taxonomy.total} confirmed IDs), level~1 items \emph{entity/loot-dependent} (\V{disc.taxonomy.entity}), and level~0 items \emph{surface primary} when a natural surface block drops them (\V{disc.taxonomy.surface_primary}) and \emph{surface derived} otherwise (\V{disc.taxonomy.surface_crafted}). Thus iron tools and glowstone dust are subsurface-dependent, books, black dye, and saddles are entity/loot-dependent, and flower dyes are surface derived. An item is \emph{gathered} if an admitted block-drop, mob-drop, or loot route attains its level, including ties with recipes or interactions, and \emph{made} otherwise; books are made, since their recipe reaches level~1 and the chest loot that also holds them is never admitted. The analysis release records the level and one attaining route for every item, and its classification script specifies the complete source lists, exclusions, interaction inputs, and parsing rules.

\input{tables/discovery-diagnostics/effects.tex}

\input{figures/discovery-diagnostics/latex_include.tex}

\input{figures/discovery-items/latex_include.tex}

\paragraph{Regime effects with uncertainty.}
Table~\ref{tab:discovery-effects} gives the matched-world effects behind Figure~\ref{fig:discovery-effects}, with 95\% intervals and world counts, and with categories ordered from low to high dependence on world knowledge; Figure~\ref{fig:discovery-diagnostics} shows the composition of breadth.
Entity/loot-dependent breadth rises in at least \V{disc.entity.pos.min} of \V{design.worlds} worlds for every model, and subsurface-dependent breadth rises with an interval that excludes zero for every model except Qwen 3.5 9B.
Surface-primary breadth rises with an interval above zero for \V{disc.primary.ci.pos} of the five models other than Qwen 3.5 9B, but by \V{disc.relgain.primary.min} to \V{disc.relgain.primary.max} of its G level; the surface-derived effect has no consistent sign across models, and \V{disc.surface.ci.incl0} of the \V{disc.surface.ci.n} surface-category intervals include zero.

\paragraph{Item-level decomposition.}
For each item $i$, let $p_{\rho,i}$ denote its confirmation rate after averaging repetitions within each world and then weighting the \V{design.worlds} common worlds equally. Since endpoint breadth counts the union of confirmed item types, the paired mean breadth difference is $\widehat\Delta_{D(T_D)}=\sum_{i}(p_{\mathrm L,i}-p_{\mathrm G,i})$. Across the \V{items.ds.n} types observed for DeepSeek, \V{items.ds.L_higher} have positive point estimates, \V{items.ds.G_higher} negative estimates, and \V{items.ds.tied} ties; positive and negative contributions sum to \V{items.ds.pos} and \V{items.ds.neg}, which recovers the net endpoint difference of \V{items.ds.net}.
Figure~\ref{fig:discovery-items} shows the largest differences: the ten largest gains are all subsurface- or entity/loot-dependent (e.g., buckets, ink sacs, iron tools, eggs), and the ten largest losses are stone and wood products made from surface materials (e.g., stone-brick and andesite slabs and stairs).

\paragraph{Behavioral measurements.}
We count the agent's own ability calls, excluding the sub-calls of composite actions and automatic observations: blind digging actions (breaking a block, digging down, tunneling), single-block inspections, and collection requests that fail because no target of the requested type is known. Table~\ref{tab:discovery-diagnostics} reports these alongside the composition measures.
In every common world under both G and L, GLM's total breadth remains below DeepSeek's, even though GLM gathers comparable numbers of subsurface-dependent items (\V{disc.subsurf.G.glm} vs.\ \V{disc.subsurf.G.ds} under G) and confirms diamonds more often (\V{disc.diamond.G.glm}/\V{disc.diamond.L.glm} vs.\ \V{disc.diamond.G.ds}/\V{disc.diamond.L.ds}); the difference lies in surface-derived items (\V{disc.derived.G.glm} vs.\ \V{disc.derived.G.ds} under G).

\input{tables/discovery-diagnostics/diagnostics.tex}

\input{tables/discovery-family-breadth/family-breadth.tex}

\paragraph{Family-merged breadth.}
Canonical IDs distinguish wood species and dye colors (e.g., \texttt{oak\_slab} and \texttt{birch\_slab}), so an agent can increase $D(T_D)$ by crafting the same recipe from different materials. We therefore recount breadth by item family, merging IDs that differ only in a wood species or a dye color (e.g., \texttt{oak\_slab} and \texttt{birch\_slab} $\to$ \texttt{*\_slab}). Table~\ref{tab:family-breadth} reports the result. Merging removes \V{fam.excess.G.ds} types from DeepSeek's G breadth, keeps the regime effect of every model except Qwen 3.5 9B positive (\V{fam.delta.five.min} to \V{fam.delta.five.max} families), leaves that of Qwen 3.5 9B near zero (\V{fam.delta.q9}, against \V{disc.delta.q9} item types), and preserves the ordering of models under both regimes.

%% file: tables/discovery-diagnostics/effects.tex
\begin{table*}[t]
\centering
\caption{Matched-world L$-$G effects in Discovery: mean over ten worlds, 95\% bootstrap interval over worlds (10{,}000 resamples), and the number of worlds with positive, zero, and negative differences. Categories are ordered from low to high dependence on world knowledge; the $D(T_D)$ column repeats Table~\ref{tab:regime-effects-full}. Averaged over all configurations, the final-breadth effect is positive in 10 of 10 worlds; over all model--world pairs, 53 of 60 are positive.}
\label{tab:discovery-effects}
\footnotesize
\setlength{\tabcolsep}{4.5pt}
\begin{tabular}{@{}lccccc@{}}
\toprule
Configuration & $D(T_D)$ & Surf.\ der. & Surf.\ prim. & Entity/loot & Subsurf. \\
\midrule
Kimi K3.1 Pro & $+35.1$ {\scriptsize[$+29.9$, $+40.5$]} & $+2.1$ {\scriptsize[$-1.3$, $+5.5$]} & $+1.6$ {\scriptsize[$-0.3$, $+3.3$]} & $+11.4$ {\scriptsize[$+9.6$, $+13.1$]} & $+20.0$ {\scriptsize[$+19.1$, $+21.1$]} \\
\quad worlds $+/0/-$ & 10/0/0 & 6/0/4 & 7/0/3 & 10/0/0 & 10/0/0 \\
DeepSeek v4.1 Flash & $+13.9$ {\scriptsize[$+5.7$, $+23.3$]} & $-4.1$ {\scriptsize[$-8.4$, $-0.1$]} & $+2.6$ {\scriptsize[$+1.3$, $+4.1$]} & $+9.3$ {\scriptsize[$+6.3$, $+12.9$]} & $+6.1$ {\scriptsize[$+3.3$, $+9.4$]} \\
\quad worlds $+/0/-$ & 9/0/1 & 3/0/7 & 10/0/0 & 10/0/0 & 10/0/0 \\
Qwen 3.8 Flash & $+19.8$ {\scriptsize[$+14.4$, $+25.8$]} & $+2.1$ {\scriptsize[$-2.6$, $+6.5$]} & $+2.0$ {\scriptsize[$+1.3$, $+2.8$]} & $+7.6$ {\scriptsize[$+6.1$, $+9.0$]} & $+8.1$ {\scriptsize[$+5.4$, $+10.8$]} \\
\quad worlds $+/0/-$ & 10/0/0 & 6/0/4 & 10/0/0 & 10/0/0 & 10/0/0 \\
GLM 5.3 Flash & $+14.5$ {\scriptsize[$+8.4$, $+20.8$]} & $+0.9$ {\scriptsize[$-3.4$, $+5.4$]} & $+1.5$ {\scriptsize[$+0.6$, $+2.3$]} & $+8.6$ {\scriptsize[$+6.1$, $+10.9$]} & $+3.5$ {\scriptsize[$+1.1$, $+6.1$]} \\
\quad worlds $+/0/-$ & 10/0/0 & 6/0/4 & 9/0/1 & 10/0/0 & 8/0/2 \\
Qwen 3.8 27B & $+20.0$ {\scriptsize[$+16.1$, $+24.2$]} & $+4.1$ {\scriptsize[$+1.0$, $+7.6$]} & $+2.2$ {\scriptsize[$+1.1$, $+3.4$]} & $+7.0$ {\scriptsize[$+5.3$, $+9.1$]} & $+6.7$ {\scriptsize[$+5.3$, $+8.0$]} \\
\quad worlds $+/0/-$ & 10/0/0 & 8/0/2 & 8/0/2 & 10/0/0 & 10/0/0 \\
Qwen 3.5 9B & $-0.6$ {\scriptsize[$-4.5$, $+4.2$]} & $-1.9$ {\scriptsize[$-3.0$, $-0.9$]} & $-1.3$ {\scriptsize[$-2.5$, $-0.02$]} & $+2.6$ {\scriptsize[$0.0$, $+5.7$]} & $0.0$ {\scriptsize[$-0.1$, $+0.2$]} \\
\quad worlds $+/0/-$ & 4/0/6 & 1/0/9 & 3/0/7 & 7/0/3 & 1/6/3 \\
\bottomrule
\end{tabular}
\end{table*}

%% file: figures/discovery-diagnostics/latex_include.tex
\begin{figure}[!htb]
\centering
\includegraphics[width=\columnwidth]{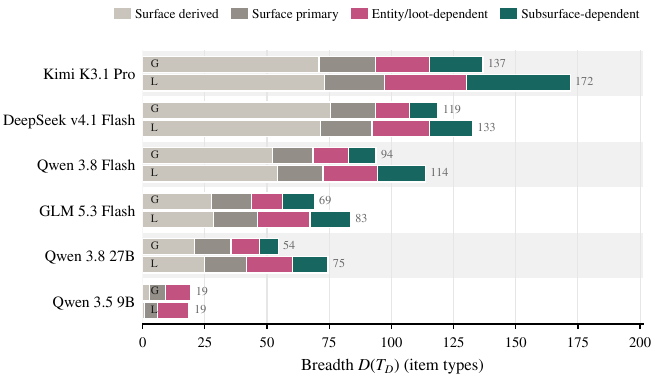}
\caption{Composition of Discovery breadth under G and L ($T_D=144{,}000$ ticks): mean breadth per run by item category, G above L for each model; numbers give $D(T_D)$. Figure~\ref{fig:discovery-effects} gives the matched-world effect on each category.}
\label{fig:discovery-diagnostics}
\end{figure}

%% file: figures/discovery-items/latex_include.tex
\begin{figure}[t]
\centering
\includegraphics[width=\columnwidth]{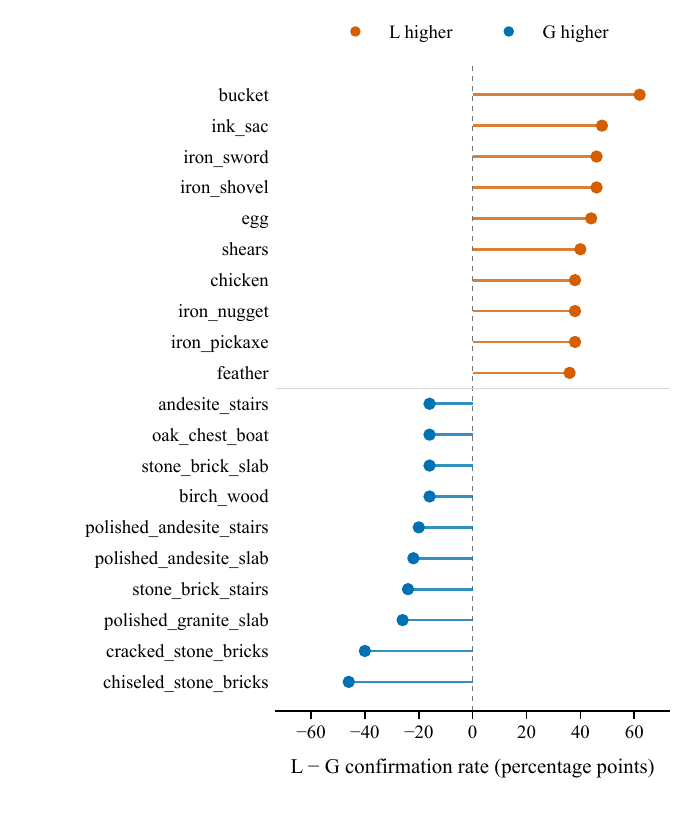}
\caption{Largest item-level differences in Discovery confirmation rate between loaded-state (L) and perception-grounded (G) knowledge for DeepSeek v4.1 Flash (144,000 simulation ticks). Rates average repetitions within each world and then weight the ten worlds equally; the 10 largest positive and negative differences are shown.}
\label{fig:discovery-items}
\end{figure}

%% file: tables/discovery-diagnostics/diagnostics.tex
\begin{table*}[t]
\centering
\caption{Discovery composition and behavior per run ($T_D=144{,}000$ simulation ticks; repetitions averaged within each world, worlds weighted equally). Dependency categories partition the 553 confirmed item IDs, ordered from low to high dependence on world knowledge: surface derived (218), surface primary (69), entity/loot-dependent (102), and subsurface-dependent (164); they sum to $D(T_D)$. \emph{Gathered} items are collected from blocks, mobs, or loot; \emph{Made} items require a recipe or tool interaction. \emph{Digging}: blind digging actions (breaking a block, digging down, tunneling); \emph{Inspect.}: single-block inspections; \emph{Unknown}: collection requests that fail because no target of the requested type is known.}
\label{tab:discovery-diagnostics}
\footnotesize
\setlength{\tabcolsep}{4.2pt}
\begin{tabular}{@{}llrrrrrrrrrr@{}}
\toprule
 & & \multicolumn{5}{c}{Breadth by dependency category} & \multicolumn{2}{c}{Source} & \multicolumn{3}{c}{Search behavior} \\
\cmidrule(lr){3-7}\cmidrule(lr){8-9}\cmidrule(l){10-12}
Configuration & & $D(T_D)$ & Surf.\ der. & Surf.\ prim. & Entity/loot & Subsurf. & Gathered & Made & Digging & Inspect. & Unknown \\
\midrule
Kimi K3.1 Pro & G & 136.80 & 70.92 & 22.62 & 21.74 & 21.52 & 41.50 & 95.30 & 113.6 & 163.3 & 27.8 \\
 & L & 171.90 & 73.00 & 24.20 & 33.14 & 41.56 & 52.30 & 119.60 & 27.3 & 12.4 & 2.6 \\
\addlinespace[2pt]
DeepSeek v4.1 Flash & G & 118.68 & 75.46 & 18.22 & 13.70 & 11.30 & 30.54 & 88.14 & 139.6 & 104.7 & 30.6 \\
 & L & 132.62 & 71.40 & 20.80 & 23.02 & 17.40 & 39.42 & 93.20 & 19.7 & 1.8 & 1.9 \\
\addlinespace[2pt]
Qwen 3.8 Flash & G & 93.84 & 52.12 & 16.34 & 14.34 & 11.04 & 29.84 & 64.00 & 140.7 & 116.6 & 57.7 \\
 & L & 113.64 & 54.20 & 18.34 & 21.94 & 19.16 & 37.26 & 76.38 & 22.6 & 4.1 & 4.9 \\
\addlinespace[2pt]
GLM 5.3 Flash & G & 69.00 & 27.70 & 16.00 & 12.58 & 12.72 & 30.60 & 38.40 & 251.2 & 381.9 & 51.4 \\
 & L & 83.48 & 28.62 & 17.46 & 21.20 & 16.20 & 38.98 & 44.50 & 47.2 & 29.5 & 6.1 \\
\addlinespace[2pt]
Qwen 3.8 27B & G & 54.48 & 20.86 & 14.66 & 11.42 & 7.54 & 27.74 & 26.74 & 175.0 & 234.7 & 21.3 \\
 & L & 74.52 & 24.96 & 16.90 & 18.46 & 14.20 & 36.88 & 37.64 & 79.6 & 38.5 & 10.7 \\
\addlinespace[2pt]
Qwen 3.5 9B & G & 19.42 & 2.64 & 6.50 & 10.08 & 0.20 & 15.86 & 3.56 & 209.5 & 2.3 & 263.7 \\
 & L & 18.82 & 0.72 & 5.24 & 12.66 & 0.20 & 17.02 & 1.80 & 9.2 & 0.2 & 1.8 \\
\bottomrule
\end{tabular}
\end{table*}

%% file: tables/discovery-family-breadth/family-breadth.tex
\begin{table*}[!t]
\centering
\caption{Discovery breadth counted by canonical item ID and by item family, where wood-species and dye-color variants of the same item are merged (rule in Appendix~\ref{app:discovery-diagnostics}). Repetitions averaged within each world, worlds weighted equally; $\Delta_{\mathrm{L}-\mathrm{G}}$ is the matched-world effect.}
\label{tab:family-breadth}
\small
\begin{tabular}{@{}lrrrrrr@{}}
\toprule
& \multicolumn{2}{c}{\textbf{G}} & \multicolumn{2}{c}{\textbf{L}} & \multicolumn{2}{c}{$\Delta_{\mathrm{L}-\mathrm{G}}$} \\
\cmidrule(lr){2-3}\cmidrule(lr){4-5}\cmidrule(l){6-7}
Configuration & IDs & Families & IDs & Families & IDs & Families \\
\midrule
Kimi K3.1 Pro & 136.80 & 116.92 & 171.90 & 147.14 & $+35.10$ & $+30.22$ \\
DeepSeek v4.1 Flash & 118.68 & 105.54 & 132.62 & 117.94 & $+13.94$ & $+12.40$ \\
Qwen 3.8 Flash & 93.84 & 82.40 & 113.64 & 101.42 & $+19.80$ & $+19.02$ \\
GLM 5.3 Flash & 69.00 & 64.00 & 83.48 & 76.32 & $+14.48$ & $+12.32$ \\
Qwen 3.8 27B & 54.48 & 50.66 & 74.52 & 67.82 & $+20.04$ & $+17.16$ \\
Qwen 3.5 9B & 19.42 & 17.74 & 18.82 & 17.84 & $-0.60$ & $+0.10$ \\
\bottomrule
\end{tabular}
\end{table*}